\pdfoutput=1
\documentclass[10pt,letterpaper]{article}

\usepackage{arsvlm_arxiv}

\papertitle{ARS-VLM: Visually Aligned Augmented Random Search\\for Vision-Language Models}

\paperauthors{%
  \textbf{Xiangwei Wang}$^{1,\dagger,\ast}$ \qquad
  \textbf{Jianyuan Lin}$^{2,\dagger}$ \qquad
  \textbf{Peng Wang}$^{3}$ \qquad
  \textbf{Saman Halgamuge}$^{1}$%
}
\affiliations{$^{1}$The University of Melbourne \qquad $^{2}$Fudan University \qquad $^{3}$Shanghai Jiao Tong University}
\authornote{$^{\dagger}$Equal contribution. \quad $^{\ast}$Correspondence to: Xiangwei Wang (\href{mailto:xiangweiw@student.unimelb.edu.au}{\texttt{xiangweiw@student.unimelb.edu.au}}).}
\institutionlogos{%
  \raisebox{-0.5\height}{\includegraphics[height=0.78cm]{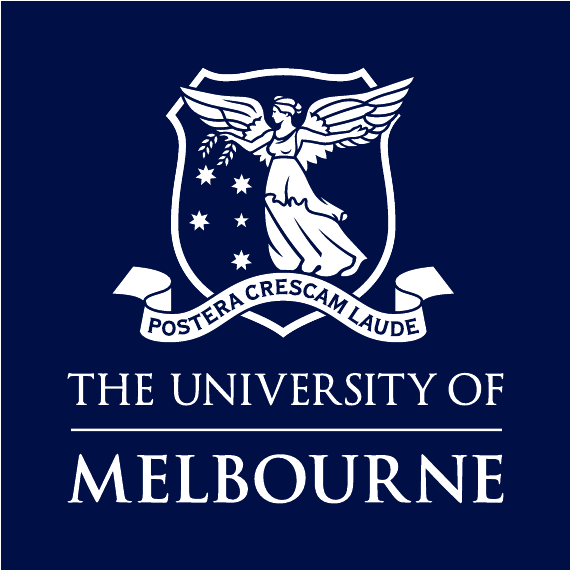}}%
  \hspace{0.65em}{\color{rulegray}\rule[-0.34cm]{0.5pt}{0.68cm}}\hspace{0.65em}%
  \raisebox{-0.5\height}{\includegraphics[height=0.78cm]{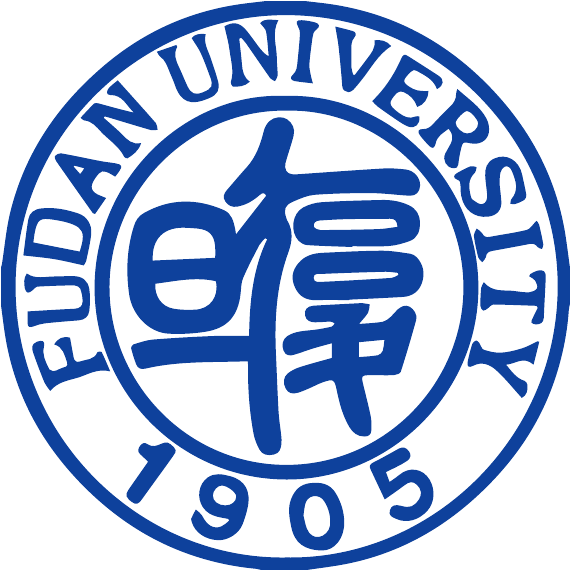}}%
  \hspace{0.65em}{\color{rulegray}\rule[-0.34cm]{0.5pt}{0.68cm}}\hspace{0.65em}%
  \raisebox{-0.5\height}{\includegraphics[height=0.78cm]{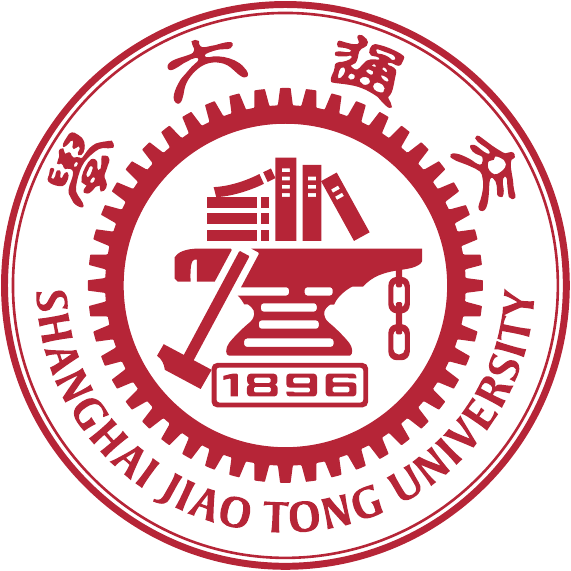}}%
}

\begin{document}

\maketitleblock

\begin{abstract}
First-order reinforcement learning, such as Group Relative Policy Optimization (GRPO), is widely used to post-train Vision-Language Models (VLMs) for multimodal reasoning. However, it incurs heavy memory and system overhead from backpropagation, optimizer states, and a reference model.
Forward-only zeroth-order (ZO) methods, such as Evolution Strategies (ES) and its top-$k$ variant Augmented Random Search (ARS), avoid these overheads via reward-guided weight perturbations, yet remain unexplored for VLM reasoning. In our study, we find that directly applying standard ES or vanilla ARS to VLMs suffers from two fundamental failure modes: (i) \emph{patch-grid overfitting}, fitting fragile grid-specific visual token artifacts while ignoring spatial invariants, and (ii) \emph{catastrophic forgetting}, degrading general perception and out-of-distribution (OOD) tasks due to unbounded high-rank parameter and policy drift. To address both issues, we propose ARS-VLM, a visually aligned Augmented Random Search framework for forward-only VLM post-training. First, ARS-VLM extracts a contrastive visual basis from bilinearly paired cross-grid views, concentrating perturbations on shared visual signals while suppressing patch-shift artifacts.
Second, ARS-VLM anchors shallow visual-fusion layers with a soft subspace prior while freeing deep reasoning layers, restraining policy drift without a reference model. Third, ARS-VLM scores perturbations on paired views to discard view-fragile directions during top-$k$ selection. We evaluate ARS-VLM on Qwen3-VL-4B and Qwen3-VL-8B across diverse multimodal reasoning tasks. On 8B, ARS-VLM reaches $61.0\%$ validation accuracy in $20$ steps versus $150$ for GRPO ($7.5\times$ fewer optimization steps) and avoids the late OOD collapse of vanilla ARS and ES. Moreover, ARS-VLM achieves $75.41\%$ single-view micro accuracy across seven OOD vision reasoning benchmarks, compared with $75.31\%$ for full-parameter GRPO, using forward-only post-training. 
\end{abstract}

\section{Introduction}
\label{sec:intro}

\begin{figure}[t]
  \centering
  \includegraphics[width=0.98\textwidth]{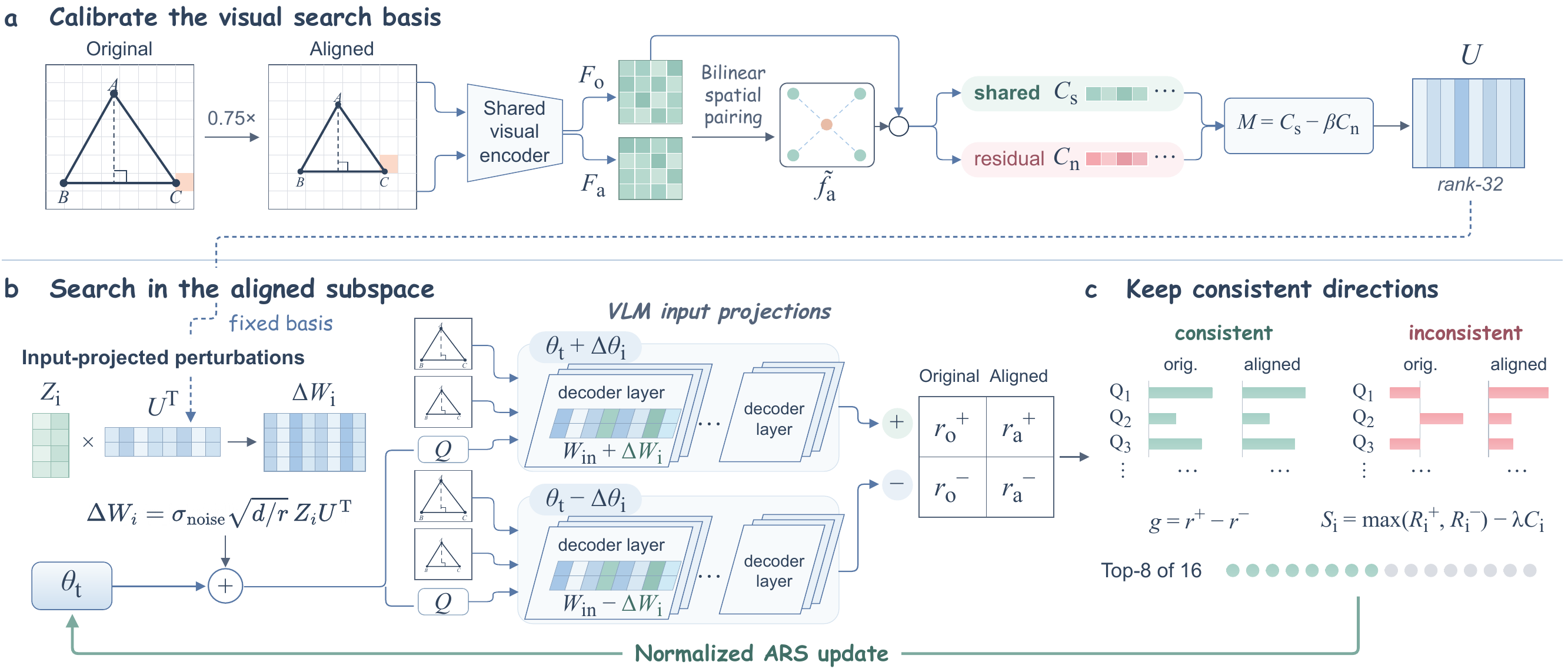}
  \caption{\textbf{Overview of \ours.}
  \textbf{(a) Visual basis:} Bilinearly paired features from original and $0.75\times$ cross-grid views yield a rank-$32$ basis $U$ from $C_{\mathrm{s}}-\beta C_{\mathrm{n}}$.
  \textbf{(b) Subspace search:} Antithetic perturbations follow $U$ during warmup, then use a layer-selective soft prior; both views are evaluated.
  \textbf{(c) Consistency selection:} The score $S_i=\max(R_i^+,R_i^-)-\lambda C_i$ selects the top-$8$ of $16$ directions.}
  \label{fig:framework}
\end{figure}

Reinforcement learning with verifiable rewards (RLVR), including Group Relative Policy Optimization (GRPO)~\cite{shao2024deepseekmath,guo2025deepseekr1,meng2025mmeureka}, improves mathematical and scientific reasoning in Vision-Language Models (VLMs). However, backpropagation through long rollouts, optimizer states, reference policies, and trainer--inference weight synchronization impose substantial memory and system costs.

Forward-only zeroth-order (ZO) methods, including MeZO~\cite{spall1992spsa,malladi2023mezo} and Evolution Strategies (ES)~\cite{salimans2017es}, estimate updates from rewards under random weight perturbations. They offer competitive text-only reasoning performance~\cite{qiu2025esscale,sun2026essam}, but unconstrained updates can cause catastrophic forgetting~\cite{abdi2026forgetting}.

We introduce Augmented Random Search (ARS)~\cite{mania2018ars}, an ES variant with top-$k$ selection, to LLM and VLM post-training and study its behavior on Qwen3-VL~\cite{qwen3vl2025} (\cref{sec:analysis}). Our analysis reveals two challenges that limit its effectiveness on VLMs. \emph{First}, image rescaling moves content across patch-grid boundaries, exposing fragile visual-token patterns that isotropic search can overfit. \emph{Second}, unconstrained updates cause parameter drift and OOD forgetting, while hard low-rank constraints across every layer limit later reasoning improvements.

Cross-grid views also provide a useful prior: they preserve diagram geometry while changing its tokenization. \ours uses this prior at three stages of forward-only search (\cref{fig:framework,sec:method}):
(i) bilinear token pairing and a contrastive visual basis determine \emph{where to perturb};
(ii) a soft subspace prior anchors shallow visual-fusion layers while releasing deeper reasoning layers; and
(iii) paired-view scoring filters inconsistent directions during top-$k$ selection.

Our contributions are:
\begin{itemize}
    \item \textbf{Empirical Findings on Forward-Only VLM Search (\cref{sec:analysis}).}
    We characterize patch-grid fragility, spectral subspace limitations, and the trade-off between full-space forgetting and deep-layer low-rank bottlenecks.

    \item \textbf{Unified Visually Aligned ARS Framework (\cref{sec:method}).}
    We combine a contrastive visual basis, a layer-selective curriculum, and dual-view selection without backpropagation or a reference model.

    \item \textbf{Comprehensive Evaluation of ARS Framework (\cref{sec:experiments}).}
    Across 4B and 8B models, \ours improves OOD retention. On 8B, it reaches the warmup target in $7.5\times$ fewer optimization steps than GRPO and achieves \textbf{$75.41\%$ single-view OOD micro accuracy} versus $75.31\%$ with validation-selected checkpoints, at $3.9\times$ lower static memory per GPU.
\end{itemize}

\section{Related Work}
\label{sec:related}

\paragraph{Reinforcement Learning for Multimodal Reasoning.}
Verifiable-reward RL via GRPO~\cite{shao2024deepseekmath,guo2025deepseekr1} has driven major gains in multimodal STEM and geometric reasoning~\cite{meng2025mmeureka,lu2024mathvista,qiao2024wemath}.
Concurrent methods such as ConsistRoll~\cite{consistroll2026} and SAGE~\cite{sage2026} add view consistency to multimodal GRPO, but operate strictly within \emph{first-order policy gradients}, which requires full backpropagation through long rollouts, AdamW states, and an in-memory reference policy $\pi_{\mathrm{ref}}$.
In contrast, \ours operates entirely in the \emph{forward-only ZO} regime, exploiting cross-grid visual geometry both to calibrate an unlabeled contrastive perturbation basis $U$ and to filter finite-difference directions without a single backward pass.

\paragraph{Zeroth-Order and Subspace Optimization at Scale.}
Derivative-free finite-difference search spans SPSA~\cite{spall1992spsa}, random gradient-free optimization~\cite{nesterov2017random}, natural ES~\cite{salimans2017es}, and ARS~\cite{mania2018ars}, which introduced antithetic top-$k$ selection and reward-dispersion normalization.
MeZO~\cite{malladi2023mezo}, Qiu~\etal~\cite{qiu2025esscale}, and ESSAM~\cite{sun2026essam} scaled seed-reconstructed perturbations~\cite{zhang2024zo_subspace} to billion-parameter LLMs; we adopt seed-regenerated inference execution as standard infrastructure.
To reduce ZO variance, Guided ES~\cite{maheswaranathan2019guided} requires surrogate gradients, while AGZO~\cite{agzo2026} uses unpaired activation PCA.
Our Unpaired PCA baseline retains $63.8\%$ of patch-shift noise. Our contrastive basis uses paired cross-view features and penalizes residual covariance, reducing shift noise and improving OOD transfer (\cref{sec:basis,sec:ablations}), following the contrastive spectral principle~\cite{abid2018contrastive}.

\paragraph{Catastrophic Forgetting and Low-Rank RL Structure.}
While first-order RLVR updates concentrate along low-rank directions~\cite{hu2022lora,schulman2025lorawithoutregret} and bound drift via $D_{\mathrm{KL}}(\pi_\theta \| \pi_{\mathrm{ref}})$, Abdi~\etal~\cite{abdi2026forgetting} showed that full-parameter ES induces dense, high-rank drift and severe catastrophic forgetting.
\ours resolves this gap in VLMs: a layer-selective soft visual prior paired with variance-floored ARS bounds shallow drift and locks Schulman policy KL~\cite{schulman2020kl} without $\pi_{\mathrm{ref}}$.

\section{Preliminaries and Empirical Analysis}
\label{sec:analysis}

\subsection{Adapting ARS to VLMs}
\label{sec:prelim}

Consider a pretrained VLM $\pi_\theta(y \mid I, q)$ parameterized by $\theta \in \mathbb{R}^P$, which maps an input image $I \in \mathbb{R}^{H\times W\times 3}$ and a text query $q$ to a reasoning trajectory $y$.
Given a training set $\mathcal{D} = \{(I_m, q_m, a_m)\}$ with a deterministic answer verifier $R(y, a) \in \{0, 1\}$, our goal is to maximize the expected verifiable reward $\mathcal{J}(\theta) = \mathbb{E}_{(I,q,a)\sim\mathcal{D},\, y\sim\pi_\theta}[R(y, a)]$ using \emph{strictly forward inference}.
To achieve this, we adapt Augmented Random Search (ARS)~\cite{mania2018ars} to VLM post-training.
At step $t$, ARS samples $N$ perturbation directions $\{\Delta\theta_i\}_{i=1}^N$ generated from random seeds, evaluates antithetic parameter pairs $\theta_t \pm \Delta\theta_i$ on a mini-batch $\mathcal{B}_t$ of $B$ problems to obtain mean rewards $(R_i^+, R_i^-)$, selects the top-$k$ directions $\mathcal{K}_t$ ranked by $\max(R_i^+, R_i^-)$, and updates
\begin{equation}
    \theta_{t+1} = \theta_t + \frac{\alpha_t}{k\,\sigma_R} \sum_{i\in\mathcal{K}_t} (R_i^+ - R_i^-) \frac{\Delta\theta_i}{\sigma_{\mathrm{noise}}},
    \label{eq:vanilla_ars}
\end{equation}
where $\sigma_{\mathrm{noise}}$ is the perturbation scale and $\sigma_R$ is the standard deviation of the $2k$ (or $2N$) evaluated rewards.
Standard full-population ES~\cite{salimans2017es,qiu2025esscale} corresponds to setting $k=N$ with $z$-scored rewards.

\subsection{Empirical Analysis on VLMs}
\label{sec:emp_analysis}

Although \cref{eq:vanilla_ars} avoids backpropagation, our empirical study on Qwen3-VL~\cite{qwen3vl2025} shows that directly applying isotropic Gaussian perturbations $\Delta\theta_i \sim \mathcal{N}(0, \sigma_{\mathrm{noise}}^2 I_P)$ to modern VLMs leads to two key observations.

\paragraph{Observation 1: Patch-Grid Fragility and Subspace Pitfalls.}
Modern dynamic-resolution VLMs~\cite{qwen3vl2025} encode an image $I$ through a $14\times 14$ ViT followed by a $2\times 2$ spatial patch merger, producing a $28\text{px}$-stride visual token grid $F_{\mathrm{o}} \in \mathbb{R}^{(G_h G_w)\times d}$ that enters the language decoder ($d=2560$ for 4B; $d=4096$ for 8B).
When we rescale diagram content by $s=0.75$ around the canvas center while preserving exact $(W, H)$ dimensions ($28\text{px}$ multiples, zero cropping), \textbf{$87.5\%$--$93.8\%$ of visual tokens shift across merger-cell boundaries} (e.g., $3{,}640 / 4{,}161$ tokens displaced on Geometry3K~\cite{lu2021intergps}), whereas mild scaling ($s=0.96$) leaves patch assignments nearly unchanged ($195.0$ vs.\ $194.0$ solved).
Meanwhile, $90.6\%$ of parameters in a $4\text{B}$--$8\text{B}$ VLM lie in the language decoder.
Isotropic perturbations therefore put over $90\%$ of search energy into text weights uncorrelated with visual tokens, causing high-variance bidirectional churn (losing $31$--$36$ previously solved problems while gaining $33$--$39$ per $20$ steps) and overfitting to fragile, grid-specific token patterns.

A natural remedy is to restrict perturbations to a low-rank visual input subspace $U \in \mathbb{R}^{d\times r}$ ($r \ll d$) calibrated from unlabeled visual features.
However, two spectral choices expose limitations (\cref{tab:probe}):
(i) \emph{Generalized eigenvalue decomposition (GEVD)}, which maximizes the Rayleigh ratio $u^\top C_{\mathrm{s}} u / (u^\top C_{\mathrm{n}} u + \epsilon \|u\|_2^2)$ between shared signal $C_{\mathrm{s}}$ and cross-view residual $C_{\mathrm{n}}$, suffers from near-zero variance amplification: because high-dimensional merger features are anisotropic ($95\%$ variance in $598$ dimensions), $(C_{\mathrm{n}}+\epsilon I)^{-1}$ strongly amplifies near-zero tail directions and retains only \textbf{$0.12\%$} of shared visual signal ($0.0008$ gradient overlap vs.\ $0.0125$ random).
(ii) \emph{Unpaired PCA} takes the top eigenvectors of the unpaired visual-feature covariance. It avoids matrix inversion and retains $73.5\%$ of shared signal, but has no explicit penalty on cross-view residual variance and retains \textbf{$63.8\%$} of patch-boundary shift noise ($\mathrm{SNR}=1.128$). On 4B it improves MMK12-Test accuracy over \ours but reduces OOD accuracy (\cref{sec:ablations}).

\paragraph{Observation 2: Full-Space Forgetting vs.\ Deep-Layer Low-Rank Bottleneck.}
Without a reference policy $\pi_{\mathrm{ref}}$, unconstrained ES and vanilla ARS accumulate dense, full-rank parameter drift across all layers. Policy KL reaches $0.0743$ for ARS at step $500$ and $0.0620$ for ES at step $1{,}000$ (\cref{fig:dynamics}c). Vanilla ARS loses $5.21$ percentage points of OOD micro accuracy from its validation-selected checkpoint to step $500$ (\cref{tab:main_8b}).
Conversely, restricting perturbations strictly to a rank-$32$ visual subspace ($\rho=1.0$) across all layers supports early warmup but hurts late-stage training: validation accuracy drops to $46.5\%$ ($93/200$) at step $425$, and $7$-OOD micro accuracy falls to $66.91\%$ at step $500$ (\cref{tab:main_8b}).
To understand this trade-off, we propagate paired cross-grid tokens through all $36$ decoder layers of Qwen3-VL.
Across the visual merger and shallow-to-middle \emph{visual-fusion layers} ($\ell \in \{0, \dots, 17\}$), cross-view visual signals stay strongly concentrated inside $\operatorname{span}(U)$ ($6.52\times$--$56.13\times$ isotropic density).
In deeper \emph{abstract reasoning layers} ($\ell \in \{18, \dots, 35\}$), however, attention and MLP blocks transform representations into symbolic coordinates where alignment with $U$ drops below $1.0\times$.
Thus, full-space updates cause shallow visual drift and forgetting, whereas hard low-rank constraints across all layers limit deep reasoning capacity.

\section{Methodology: ARS-VLM Framework}
\label{sec:method}

Guided by Observations 1 and 2, we propose \ours (\cref{fig:framework}), which combines three VLM-tailored components: a bilinearly aligned contrastive visual basis (\cref{sec:basis}), a layer-selective soft subspace curriculum (\cref{sec:curriculum}), and dual-view consistency selection (\cref{sec:selection}).

\subsection{Bilinearly Aligned Contrastive Visual Basis}
\label{sec:basis}

\ours addresses both subspace pitfalls in Observation 1 (\cref{sec:emp_analysis}) through continuous spatial pairing and an inversion-free contrastive operator (\cref{fig:framework}a), calibrated once in ${\sim}30\text{s}$ on $M_{\mathrm{cal}}=32$ unlabeled training images with zero reward queries.

\paragraph{Deterministic Cross-Grid View and Bilinear Spatial Pairing.}
Given an unlabeled calibration image $I_{\mathrm{o}} \in \mathbb{R}^{H\times W\times 3}$ (padded or resized to $28\text{px}$ multiples so $G_h = H/28, G_w = W/28$), we construct a deterministically aligned view $I_{\mathrm{a}}$ on an identical white canvas $(W, H)$ by scaling the diagram by $s=0.75$ around the center with offset $(o_x, o_y) = (\lfloor(W - sW)/2\rfloor, \lfloor(H - sH)/2\rfloor)$ and mild photometric jitter ($1.08\times$ brightness/contrast).
Passing $I_{\mathrm{o}}$ and $I_{\mathrm{a}}$ through the frozen visual tower and patch merger produces token grids $F_{\mathrm{o}}, F_{\mathrm{a}} \in \mathbb{R}^{G_h \times G_w \times d}$.

Rather than matching discrete token indices, which incurs quantization error across $28\text{px}$ cell borders, we map the continuous center $((c+0.5)\frac{W}{G_w},\, (r'+0.5)\frac{H}{G_h})$ of each original token $t=(r', c)$ onto exact continuous sub-cell coordinates $(u_t, v_t)$ in the aligned grid:
\begin{equation}
    u_t = \frac{o_x}{W} G_w + s\big(c + \tfrac{1}{2}\big) - \tfrac{1}{2}, \quad
    v_t = \frac{o_y}{H} G_h + s\big(r' + \tfrac{1}{2}\big) - \tfrac{1}{2}.
    \label{eq:subcell_coords}
\end{equation}
We reconstruct the spatially paired feature $\tilde{f}_{\mathrm{a}, t} = \mathcal{I}_{\mathrm{bilin}}(F_{\mathrm{a}}; u_t, v_t) \in \mathbb{R}^d$ via 2D bilinear interpolation (\texttt{grid\_sample}).
On Qwen3-VL, \cref{eq:subcell_coords} raises cross-view token cosine similarity from $0.720 \to 0.769$ and cuts total shift-noise trace $\Tr(C_{\mathrm{n}})$ by \textbf{$29.8\%$} ($333.1 \to 234.0$) before eigendecomposition.

\paragraph{Contrastive Cross-View Difference Operator.}
For each non-background token pair $(f_{\mathrm{o}, t}, \tilde{f}_{\mathrm{a}, t})$ (after subtracting the global mean), we decompose the cross-grid representation into its view-invariant mean and view-shift residual:
\begin{equation}
    s_t = \tfrac{1}{2}\big(f_{\mathrm{o}, t} + \tilde{f}_{\mathrm{a}, t}\big), \qquad
    n_t = f_{\mathrm{o}, t} - \tilde{f}_{\mathrm{a}, t},
    \label{eq:sn_decomp}
\end{equation}
forming empirical second-moment matrices $C_{\mathrm{s}} = \frac{1}{T_{\mathrm{tok}}}\sum_t s_t s_t^\top$ and $C_{\mathrm{n}} = \frac{1}{T_{\mathrm{tok}}}\sum_t n_t n_t^\top$.
Following contrastive spectral analysis~\cite{abid2018contrastive}, we extract the top-$r$ ($r=32$) unit-norm eigenvectors $U = [u_1, \dots, u_r] \in \mathbb{R}^{d\times r}$ ($U^\top U = I_r$) of the symmetric difference operator:
\begin{equation}
    M_\beta = C_{\mathrm{s}} - \beta\, C_{\mathrm{n}}, \qquad \beta > 0 \quad (\text{default } \beta = 0.35).
    \label{eq:contrastive_op}
\end{equation}

\begin{proposition}[Cross-Covariance Identity and Directional SNR Amplification]
\label{prop:contrastive}
Let $C_{\mathrm{pool}} = \frac{1}{2}\mathbb{E}[f_{\mathrm{o}} f_{\mathrm{o}}^\top + \tilde{f}_{\mathrm{a}} \tilde{f}_{\mathrm{a}}^\top]$ denote the unpaired pooled covariance and $C_{\mathrm{cross}} = \frac{1}{2}\mathbb{E}[f_{\mathrm{o}}\tilde{f}_{\mathrm{a}}^\top + \tilde{f}_{\mathrm{a}} f_{\mathrm{o}}^\top]$ the symmetric cross-view covariance. Then:
\begin{enumerate}
    \item \textbf{Inversion-free cross-view filtering:} $M_\beta$ admits the exact decomposition
    \begin{equation}
        M_\beta = \big(\tfrac{1}{2} - 2\beta\big) C_{\mathrm{pool}} + \big(\tfrac{1}{2} + 2\beta\big) C_{\mathrm{cross}}.
        \label{eq:cross_cov_id}
    \end{equation}
    Hence for $\beta > \frac{1}{4}$ (e.g., $\beta=0.35$, giving $M_{0.35} = 1.2\,C_{\mathrm{cross}} - 0.2\,C_{\mathrm{pool}}$), maximizing $\Tr(U^\top M_\beta U)$ strictly rewards positive cross-view feature agreement while penalizing view-specific variance, without inverting $C_{\mathrm{n}}$ or entering its near-zero variance directions.
    \item \textbf{Norm conservation and $d/r$ directional gain:} For any decoder input projection $W \in \mathbb{R}^{d_{\mathrm{out}}\times d}$ and $Z \sim \mathcal{N}(0, I_{d_{\mathrm{out}}\times r})$, the right-projected perturbation $\Delta W = \sigma_{\mathrm{noise}}\sqrt{d/r}\, Z U^\top$ preserves exact isotropic Frobenius energy $\mathbb{E}\|\Delta W\|_F^2 = \sigma_{\mathrm{noise}}^2 d_{\mathrm{out}} d$, while amplifying the expected squared directional derivative $\mathbb{E}[\langle \nabla_W \mathcal{J}, \Delta W\rangle^2]$ along the $U$-aligned visual gradient component $\nabla_W \mathcal{J}\, U U^\top$ by a factor of $d/r$ ($128\times$ for $d=4096, r=32$).
\end{enumerate}
\end{proposition}

A complete proof is provided in Appendix~\ref{sec:supp_proof}.
Our Unpaired PCA baseline uses the leading eigenvectors of the unpaired visual-feature covariance. It differs from setting $\beta=0$ in our operator, which retains paired-mean covariance $C_{\mathrm{s}}=\tfrac12(C_{\mathrm{pool}}+C_{\mathrm{cross}})$. The comparison therefore evaluates the paired contrastive construction as a whole.
Empirically on Qwen3-VL (\cref{tab:probe}), $U$ ($r=32$, only $0.78\%$--$1.25\%$ of ambient dimension $d$) retains \textbf{$70.16\%$} of shared visual signal ($56.1\times$ isotropic density), suppresses $31.1\%$ of grid-shift noise relative to PCA ($\mathrm{SNR}=1.562$ vs.\ $1.128$), and increases single-batch reward directional derivative magnitude $|\langle \nabla\mathcal{J}, \Delta\theta\rangle|$ by \textbf{$+37.1\%$} ($0.0938$ vs.\ $0.0684$).

\subsection{Layer-Selective Soft Subspace Curriculum}
\label{sec:curriculum}

Motivated by the layer-wise specialization in Observation 2, \ours applies a brief low-rank warmup across all layers for fast initial alignment, and then keeps a soft visual subspace prior exclusively in shallow visual-fusion layers while freeing deep reasoning layers (\cref{fig:framework}b).

\paragraph{Two-Stage Norm-Preserving Perturbation Schedule.}
For each 2D decoder input projection $W_\ell \in \mathbb{R}^{d_{\mathrm{out}}\times d}$ ($W_q, W_k, W_v, W_{\mathrm{gate}}, W_{\mathrm{up}}$) at layer $\ell \in \{0,\dots,L-1\}$, \ours defines a layer- and time-dependent mixing weight $\rho_{t,\ell} \in [0, 1]$:
\begin{equation}
    \rho_{t,\ell} =
    \begin{cases}
        1, & 1 \le t \le T_{\mathrm{w}}, \\
        \rho_{\mathrm{soft}}, & t > T_{\mathrm{w}} \text{ and } \ell < L_{\mathrm{vf}}, \\
        0, & t > T_{\mathrm{w}} \text{ and } \ell \ge L_{\mathrm{vf}},
    \end{cases}
    \label{eq:rho_schedule}
\end{equation}
with warmup horizon $T_{\mathrm{w}} = 20$, visual-fusion boundary $L_{\mathrm{vf}} = 18$, and $\rho_{\mathrm{soft}} = 0.25$.
Sampling independent standard Gaussian matrices $Z_{\mathrm{full}, i,\ell} \in \mathbb{R}^{d_{\mathrm{out}}\times d}$ and $Z_{\mathrm{sub}, i,\ell} \in \mathbb{R}^{d_{\mathrm{out}}\times r}$ from deterministic tensor seeds, we form:
\begin{equation}
    \Delta W_{i,\ell} = \sigma_{\mathrm{noise}} \!\left( \sqrt{1 - \rho_{t,\ell}^2}\, Z_{\mathrm{full}, i,\ell} + \rho_{t,\ell}\sqrt{\frac{d}{r}}\, Z_{\mathrm{sub}, i,\ell} U^\top \right)\!.
    \label{eq:soft_perturb}
\end{equation}
By orthogonality $U^\top U = I_r$, \cref{eq:soft_perturb} satisfies $\mathbb{E}\|\Delta W_{i,\ell}\|_F^2 = \sigma_{\mathrm{noise}}^2 d_{\mathrm{out}} d$ for every $\rho_{t,\ell}\in[0,1]$, preserving expected perturbation energy across the two stages. In shallow visual-fusion layers, the subspace term contributes $\rho_{\mathrm{soft}}^2 d/r$ times the isotropic per-direction variance: $5\times$ on 4B and $8\times$ on 8B. Including the full-space term gives total variance factors of $5.9375$ and $8.9375$ along unit directions in $\operatorname{span}(U)$.

\paragraph{Cosine Annealing and Variance-Floored Step Normalization.}
In long-horizon RLVR with binary rewards, rising accuracy causes the top-$k$ reward standard deviation $\sigma_R$ in \cref{eq:vanilla_ars} which shrinks toward zero, which unintentionally amplifies raw parameter steps $1/\sigma_R$ late in training and drives unbounded Frobenius drift (\cref{fig:dynamics}c).
In Stage 2 ($t > T_{\mathrm{w}}$), we pair \cref{eq:soft_perturb} with cosine step-size annealing $\alpha_t = \alpha_0 \big(0.25 + 0.375(1 + \cos(\pi \frac{t - T_{\mathrm{w}}}{T - T_{\mathrm{w}}}))\big)$ and an effective reward dispersion floor $\sigma_{\mathrm{eff}} = \max(\sigma_{\mathrm{all}}, \sigma_{\mathrm{sel}})$ taking the maximum dispersion across all $2N$ and selected $2k$ evaluations.
Together with $\rho_{\mathrm{soft}}$ in shallow layers, this forms an \textbf{implicit geometric trust region} without storing a reference model $\pi_{\mathrm{ref}}$.

\subsection{Dual-View Consistency Selection and Execution}
\label{sec:selection}

Subspace projection shapes \emph{where} candidate perturbations point, while our third component governs \emph{which} perturbations are selected (\cref{fig:framework}c).
For each mini-batch $\mathcal{B}_t = \{b\}_{b=1}^B$, every antithetic candidate $\theta_t \pm \Delta\theta_i$ ($i=1,\dots,N$) is evaluated in a single batched \texttt{vLLM} pass across both the original diagram $I_{b,\mathrm{o}}$ and its deterministic $0.75\times$ cross-grid view $I_{b,\mathrm{a}}$, yielding binary verifiable rewards $(r_{i,b,\mathrm{o}}^\pm, r_{i,b,\mathrm{a}}^\pm)$.

Let $R_i^\pm = \frac{1}{2B}\sum_{b=1}^B (r_{i,b,\mathrm{o}}^\pm + r_{i,b,\mathrm{a}}^\pm)$ be the pooled dual-view reward and define the per-direction \emph{cross-view directional inconsistency}:
\begin{equation}
    C_i = \frac{1}{B} \sum_{b=1}^B \Big| \big(r_{i,b,\mathrm{o}}^+ - r_{i,b,\mathrm{o}}^-\big) - \big(r_{i,b,\mathrm{a}}^+ - r_{i,b,\mathrm{a}}^-\big) \Big|.
    \label{eq:inconsistency}
\end{equation}
Notice that $C_i = 0$ if and only if the finite-difference sign and magnitude $r^+ - r^-$ agree on every problem across both patch-grid discretizations, whereas directions exploiting fragile cell-border token patterns incur large $C_i$.
We rank all $N$ directions by the consistency-penalized score:
\begin{equation}
    S_i = \max(R_i^+, R_i^-) - \lambda\, C_i \qquad (\lambda = 0.1),
    \label{eq:selection_score}
\end{equation}
select the top-$k$ indices $\mathcal{K}_t = \operatorname{TopK}(\{S_i\}_{i=1}^N, k)$, and apply the normalized update:
\begin{equation}
    \theta_{t+1} = \theta_t + \frac{\alpha_t}{k\,\sigma_{\mathrm{eff}}} \sum_{i\in\mathcal{K}_t} (R_i^+ - R_i^-) \frac{\Delta\theta_i}{\sigma_{\mathrm{noise}}}.
    \label{eq:ours_update}
\end{equation}

\begin{algorithm}[t]
\caption{\ours: Forward-Only Visually Aligned ARS}
\label{alg:ars_vlm}
\small
\begin{algorithmic}[1]
\Require Pretrained VLM $\theta_0$, unlabeled images $\{I_m\}_{m=1}^{32}$, rank $r=32$, $\beta=0.35$, $N=16, k=8$, $\sigma_{\mathrm{noise}}=10^{-3}$, $\alpha_0=10^{-4}$, $\lambda=0.1$.
\State \textbf{Basis Calibration (${\sim}30\text{s}$, once):} Extract merger grids $(F_{\mathrm{o}}, F_{\mathrm{a}})$ at $s=0.75$, pair $\tilde{f}_{\mathrm{a}}$ via bilinear \cref{eq:subcell_coords}, compute $C_{\mathrm{s}}, C_{\mathrm{n}}$ (\cref{eq:sn_decomp}), and set $U \leftarrow \operatorname{TopEig}_r(C_{\mathrm{s}} - \beta C_{\mathrm{n}})$.
\For{step $t = 1, \dots, T$}
    \State Sample batch $\mathcal{B}_t$ of $B$ problems with paired views $(I_{b,\mathrm{o}}, I_{b,\mathrm{a}})$.
    \State Set layer schedule $\rho_{t,\ell}$ (\cref{eq:rho_schedule}) and step size $\alpha_t$.
    \For{direction $i = 1, \dots, N$ and sign $\xi \in \{+1, -1\}$ \textbf{in parallel}}
        \State Perturb $\theta_t + \xi\Delta\theta_i$ in-place from seed $(t,i)$ via \cref{eq:soft_perturb}.
        \State Run forward rollout on $\{(I_{b,\mathrm{o}}, q_b), (I_{b,\mathrm{a}}, q_b)\}_{b\in\mathcal{B}_t}$; restore $\theta_t$.
    \EndFor
    \State Compute $R_i^\pm$, cross-view inconsistency $C_i$ (\cref{eq:inconsistency}), and $S_i$ (\cref{eq:selection_score}).
    \State Select $\mathcal{K}_t = \operatorname{TopK}(S, k)$, compute $\sigma_{\mathrm{eff}} = \max(\sigma_{\mathrm{all}}, \sigma_{\mathrm{sel}})$, and accumulate \cref{eq:ours_update} tensor-wise in \texttt{fp32} before casting to \texttt{bf16}.
\EndFor
\end{algorithmic}
\end{algorithm}

\paragraph{Zero-Overhead Inference-Engine Realization.}
\cref{alg:ars_vlm} runs entirely inside persistent \texttt{vLLM}~\cite{kwon2023vllm} workers ($U$ takes $<0.5\text{ MB}$).
Workers regenerate $\Delta W_{i,\ell}$ in-place from 64-bit seeds, clear multimodal prefix caches, run PagedAttention decoding, and accumulate \cref{eq:ours_update} tensor-wise in \texttt{fp32} before casting to \texttt{bf16}. No backward activations, gradients, optimizer states, or reference-model copy are stored; forward activations, KV caches, and temporary update buffers are still needed during execution.

\section{Experiments}
\label{sec:experiments}

\begin{table}[t]
  \centering
  \caption{\textbf{Multimodal STEM and OOD generalization results for Qwen3-VL-8B-Instruct post-training and public RLVR checkpoints.} Our 8B runs train on MMK12 for $500$ steps; all evaluations use greedy decoding with at most $4{,}096$ tokens. Accuracy ($\%$) on MMK12-Test ($2{,}000$ items) and seven OOD benchmarks ($7{,}610$ items). \textbf{FO}: first-order training; \textbf{ZO}: forward-only training. ``Best'' denotes selection on MMK12-Dev, not on the test benchmarks. Best single-view ZO in \textbf{bold}. PCA Micro is recomputed from the displayed per-benchmark accuracies; its We-Math and Geo3K entries are shown to one decimal place.}
  \label{tab:main_8b}
  \footnotesize
  \setlength{\tabcolsep}{4.0pt}
  \renewcommand{\arraystretch}{1.10}
  \begin{adjustbox}{max width=\textwidth}
  \begin{tabular}{l c cc ccccccc cc}
    \toprule
    \multirow{2}{*}{\textbf{Method}} & \multirow{2}{*}{\textbf{Mode}} & \multicolumn{2}{c}{\textbf{MMK12-Test} (ID)} & \multicolumn{7}{c}{\textbf{Seven Out-of-Distribution Vision Reasoning Benchmarks}} & \multicolumn{2}{c}{\textbf{7-OOD Summary}} \\
    \cmidrule(lr){3-4} \cmidrule(lr){5-11} \cmidrule(lr){12-13}
    & & All & Math & MathVista & We-Math & SciQA & MMStar & Geo3K & MathVis & LogicVis & \textbf{Macro} & \textbf{Micro} \\
    \midrule
    Pretrained Base (Step 0) & -- & 49.20 & 30.00 & 74.70 & 70.52 & 93.60 & 69.27 & 54.74 & 33.55 & 50.00 & 63.77 & 73.01 \\
    \midrule
    \multicolumn{13}{l}{\emph{First-Order RL: Public Checkpoints and Our Full-Parameter GRPO Baseline}} \\
    MM-Eureka-Qwen-7B~\cite{meng2025mmeureka} & FO & 57.65 & 47.40 & 69.10 & 63.33 & 89.14 & 63.80 & 40.77 & 32.24 & 44.87 & 57.61 & 66.91 \\
    PRISM-DAPO-8B~\cite{wang2026prism} & FO & 7.20 & 16.40 & 43.60 & 28.62 & 79.33 & 51.47 & 42.26 & 20.39 & 16.52 & 40.31 & 48.57 \\
    Full-Param GRPO (Best)~\cite{shao2024deepseekmath} & FO & 55.15 & 29.80 & 76.30 & 73.97 & 94.25 & 70.13 & 58.90 & 39.80 & 56.47 & 67.12 & 75.31 \\
    Full-Param GRPO (Step 500)~\cite{shao2024deepseekmath} & FO & 53.25 & 26.60 & 77.80 & 74.71 & 94.05 & 69.93 & 58.57 & 41.12 & 55.13 & 67.33 & 75.53 \\
    \midrule
    \multicolumn{13}{l}{\emph{Forward-Only Zeroth-Order Baselines \& Subspace Ablations (Zero Gradients, Zero Optimizer States, No $\pi_{\mathrm{ref}}$)}} \\
    MeZO-RGE ($N=16$, Step 500)~\cite{malladi2023mezo} & ZO & 49.75 & 29.80 & 75.00 & 71.03 & 93.90 & 69.40 & 56.74 & 33.55 & 51.56 & 64.46 & 73.52 \\
    Vanilla ES (Best)~\cite{qiu2025esscale} & ZO & 48.25 & 22.00 & 73.80 & 73.33 & 93.16 & 69.13 & 55.91 & 35.20 & 50.22 & 64.39 & 73.56 \\
    Vanilla ES (Step 500)~\cite{qiu2025esscale} & ZO & 47.90 & 24.00 & 72.50 & 73.28 & 93.16 & 66.80 & 56.74 & 33.88 & 52.01 & 64.05 & 73.04 \\
    Vanilla ARS (Best, Step 90)~\cite{mania2018ars} & ZO & 53.30 & \textbf{36.60} & 75.70 & 72.13 & 94.15 & 68.47 & 56.91 & 34.87 & 51.79 & 64.86 & 73.82 \\
    Vanilla ARS (Step 500)~\cite{mania2018ars} & ZO & 49.90 & 33.20 & 65.10 & 69.77 & 88.55 & 63.07 & 51.58 & 29.93 & 49.78 & 59.68 & 68.61 \\
    Hard Subspace ($\rho=1.0$, Step 500) & ZO & 45.35 & 26.80 & 63.20 & 66.67 & 87.01 & 63.53 & 44.26 & 30.59 & 52.01 & 58.18 & 66.91 \\
    Unpaired PCA & ZO & 51.20 & 28.40 & 74.80 & 73.9 & 94.35 & 68.60 & 56.4 & 37.50 & 52.90 & 65.49 & 74.32 \\
    Random Orthonormal Subspace (Best) & ZO & 49.10 & 26.40 & 75.50 & 73.74 & 94.79 & 68.00 & \textbf{57.57} & 35.53 & 49.33 & 64.92 & 74.18 \\
    \rowcolor{oursrow}
    \ours \textbf{(Ours, Best)} & \textbf{ZO} & 53.70 & 30.40 & \textbf{75.80} & 75.17 & \textbf{95.04} & \textbf{70.00} & 56.07 & \textbf{41.45} & \textbf{54.24} & \textbf{66.83} & \textbf{75.41} \\
    \rowcolor{oursrow}
    \ours \textbf{(Ours, Step 500)} & \textbf{ZO} & \textbf{54.20} & 31.60 & 75.20 & \textbf{75.52} & 94.60 & 69.87 & \textbf{57.57} & 36.84 & 52.68 & 66.04 & 75.11 \\
    \bottomrule
  \end{tabular}
  \end{adjustbox}
\end{table}

\subsection{Experimental Setup}
\label{sec:setup}

\paragraph{Models and Benchmarks.}
We evaluate \ours on \textbf{Qwen3-VL-8B-Instruct} ($d=4096$, $36$ layers) and \textbf{Qwen3-VL-4B-Instruct} ($d=2560$, $36$ layers)~\cite{qwen3vl2025} across three suites:
(1) \textbf{Multimodal STEM \& 7-Benchmark OOD:} Post-training on \textbf{MMK12}~\cite{meng2025mmeureka} ($B=16$; $500$ training steps for both 4B and 8B, including all ablations), validated on $200$-item Dev and tested on held-out $2{,}000$-item \textbf{MMK12-Test} ($500$ each in Math, Physics, Chemistry, Biology) plus \textbf{seven OOD vision benchmarks ($7{,}610$ items)}: MathVista ($1{,}000$)~\cite{lu2024mathvista}, We-Math ($1{,}740$)~\cite{qiao2024wemath}, ScienceQA-IMG ($2{,}017$)~\cite{lu2022scienceqa}, MMStar ($1{,}500$)~\cite{chen2024mmstar}, Geometry3K ($601$)~\cite{lu2021intergps}, MathVision ($304$)~\cite{wang2024mathvision}, and LogicVista ($448$)~\cite{xiao2024logicvista}.
(2) \textbf{16-Run Forgetting Suite:} Following Abdi~\etal~\cite{abdi2026forgetting}, $150$-step Qwen3-VL-4B RL across Countdown, GSM8K~\cite{cobbe2021gsm8k}, MATH-500~\cite{hendrycks2021math}, and OlympiadBench~\cite{he2024olympiadbench}, tracking HellaSwag~\cite{zellers2019hellaswag} and MMStar~\cite{chen2024mmstar} every $10$ steps.
(3) \textbf{Subspace \& View Probe:} Multi-seed Geometry3K~\cite{lu2021intergps} spectral diagnostics.

\begin{table}[t]
  \centering
  \caption{\textbf{Cross-scale results and component ablations on Qwen3-VL-4B-Instruct} (MMK12 post-training, $500$ steps for every method and ablation; Dev peak and best-validation Test / 4-OOD accuracy across $5{,}858$ OOD items). All four ablation variants fall below the pretrained base on OOD micro accuracy ($71.77\%$). Parenthesized differences are percentage points relative to the full method. The single-view variant removes both the second view and its consistency penalty.}
  \label{tab:ablations_4b}
  \footnotesize
  \setlength{\tabcolsep}{4.8pt}
  \renewcommand{\arraystretch}{1.12}
  \begin{adjustbox}{max width=\textwidth}
  \begin{tabular}{l c cc cccc c}
    \toprule
    \textbf{Method / Ablation Variant} & \textbf{Mode} & \textbf{Dev Peak} & \textbf{MMK12-Test} & \textbf{We-Math} & \textbf{SciQA} & \textbf{MMStar} & \textbf{Geo3K} & \textbf{4-OOD Micro} \\
    \midrule
    Pretrained Base (4B, Step 0) & -- & 39.00 & 38.35 & 61.09 & 92.12 & \textbf{65.73} & 49.42 & 71.77 \\
    \midrule
    Full-Param GRPO~\cite{shao2024deepseekmath} & FO & 49.50 & 41.95 & 63.85 & 92.02 & 65.13 & \textbf{49.92} & 72.45 \\
    \midrule
    MeZO-RGE ($N=16$, Antithetic SPSA)~\cite{malladi2023mezo} & ZO & 44.50 & 38.50 & 56.44 & 90.28 & 63.73 & 42.93 & 68.57 \\
    Vanilla ARS~\cite{mania2018ars} & ZO & 48.50 & 40.60 & 62.93 & 92.27 & \textbf{65.73} & 48.42 & 72.26 \\
    Vanilla ES~\cite{qiu2025esscale} & ZO & \textbf{51.00} & 40.30 & \textbf{66.95} & 91.72 & 65.00 & 49.75 & 73.22 \\
    \rowcolor{oursrow}
    \ours \textbf{(Full Method)} & \textbf{ZO} & 50.00 & 40.10 & 66.55 & \textbf{92.66} & 65.33 & 48.92 & \textbf{73.42} \\
    \midrule
    \multicolumn{9}{l}{\emph{Controlled Component Ablations of \ours (500 Training Steps, Qwen3-VL-4B, Best-Dev Checkpoints)}} \\
    \quad w/ Unpaired PCA & ZO & 47.00 \drop{-3.00} & \textbf{42.25} & 62.76 & 91.77 & 64.13 & 47.09 & 71.49 \drop{-1.93} \\
    \quad w/ Random Orthonormal Subspace ($U_{\mathrm{rand}}$) & ZO & 47.50 \drop{-2.50} & 39.50 & 61.38 & 92.51 & 64.80 & 47.75 & 71.58 \drop{-1.84} \\
    \quad w/o Dual-View Scoring (Single-View, $\lambda=0$) & ZO & 45.50 \drop{-4.50} & 38.70 & 62.64 & 91.22 & 64.47 & 47.59 & 71.41 \drop{-2.01} \\
    \quad w/o Soft Prior (Hard Subspace $\rho = 1.0$) & ZO & 48.50 \drop{-1.50} & 39.10 & 61.32 & 90.73 & 63.67 & 45.76 & 70.45 \drop{-2.97} \\
    \bottomrule
  \end{tabular}
  \end{adjustbox}
\end{table}

\paragraph{Baselines and Implementation.}
We compare the following methods under the stated training horizons; rollout counts per optimization step differ:
(i) \textbf{Full-Param GRPO}~\cite{shao2024deepseekmath} ($G=8$, $\text{lr}=10^{-6}$, AdamW, frozen $\pi_{\mathrm{ref}}$ with $\beta_{\mathrm{KL}}=0.01$);
(ii) \textbf{MeZO-RGE}~\cite{malladi2023mezo} ($N=16$ antithetic SPSA pairs, $\sigma_{\mathrm{noise}}=10^{-3}$, $\eta=10^{-6}$);
(iii) \textbf{Vanilla ES}~\cite{salimans2017es,qiu2025esscale} ($P=30$, $\sigma=10^{-3}$, $\alpha=5\times 10^{-4}$);
(iv) \textbf{Vanilla ARS}~\cite{mania2018ars} ($N=16, k=8, \sigma_{\mathrm{noise}}=10^{-3}, \alpha=10^{-4}$); and
(v) public RLVR models \textbf{MM-Eureka-Qwen-7B}~\cite{meng2025mmeureka} and \textbf{PRISM-DAPO-8B}~\cite{wang2026prism}.
\ours calibrates $U$ on $32$ unlabeled images ($s=0.75, r=32, \beta=0.35$) in ${\sim}30\text{s}$ with $T_{\mathrm{w}}=20$, $\rho_{\mathrm{soft}}=0.25$ ($\ell < 18$), and $\lambda=0.1$.
Evaluations use greedy decoding (\texttt{temperature=0}; max $4{,}096$ tokens in \cref{tab:main_8b}, $2{,}048$ in \cref{tab:ablations_4b}) in \texttt{vLLM}~\cite{kwon2023vllm}.
At $B=16$, GRPO uses $BG=128$ trajectories per step; single-view ARS and MeZO-RGE use $2NB=512$, and dual-view \ours uses $4NB=1{,}024$. These comparisons are not matched for total rollout count. Micro accuracy weights each benchmark by its test-set size; macro accuracy averages benchmark accuracies equally. Accuracy differences below are in percentage points (pp).

\subsection{Main Multimodal STEM and OOD Results}
\label{sec:main_results}

\begin{figure}[t]
  \centering
  \includegraphics[width=\textwidth]{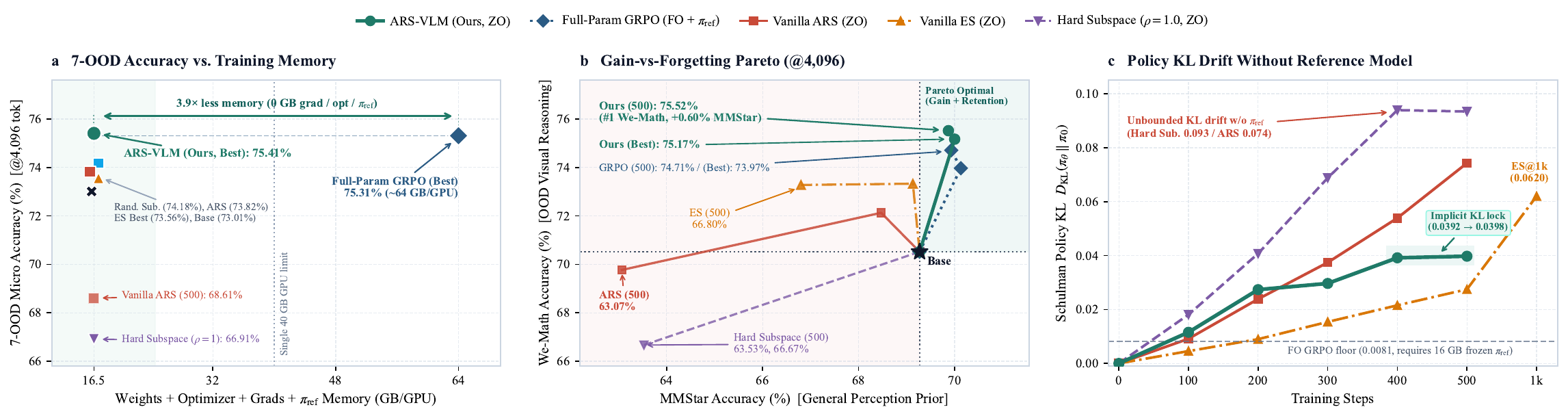}
  \caption{\textbf{Static memory, OOD retention, and policy drift on Qwen3-VL-8B.}
  \textbf{(a)} With validation-selected checkpoints, \ours achieves $75.41\%$ $7$-OOD micro accuracy at ${\sim}16.5\text{ GB}$ of static memory, versus GRPO's $75.31\%$ at ${\sim}64\text{ GB/GPU}$ ($3.9\times$ ratio). Both figures exclude runtime activations, KV caches, and temporary buffers; neither is peak training memory.
  \textbf{(b)} At step $500$, \ours improves both We-Math ($75.52\%$) and MMStar ($+0.60$ pp over Base); GRPO achieves $74.71\%$ on We-Math. Evaluation uses a $4{,}096$-token limit.
  \textbf{(c)} Reference-free KL dynamics: \ours stabilizes Schulman KL ($0.0392\to 0.0398$) without $\pi_{\mathrm{ref}}$, unlike the linear drift in ARS ($0.0743$) and Hard Subspace ($0.0934$), or ES ($0.0620$ at $1\text{k}$).}
  \label{fig:dynamics}
\end{figure}

\paragraph{Qwen3-VL-8B Results (\cref{tab:main_8b}).}
Vanilla ARS loses OOD accuracy late in training: micro accuracy falls from $73.82\%$ at its validation-selected checkpoint to $68.61\%$ at step $500$, below the pretrained base. Hard subspace constraints also degrade performance, while \ours retains gains over Base at step $500$.

With validation-selected checkpoints, \ours reaches $75.41\%$ single-view OOD micro accuracy versus GRPO's $75.31\%$, although GRPO has higher macro and in-domain accuracy. At step $500$, \ours has higher MMK12-Test accuracy ($54.20\%$ vs.\ $53.25\%$), while GRPO has higher OOD micro accuracy. Thus, the advantage depends on the metric and checkpoint. Label-free dual-view fallback yields comparable OOD results for both methods at $4{,}096$ tokens; detailed results are in Appendix~\ref{sec:supp_dv}.

\subsection{Cross-Scale Validation and Ablation Studies}
\label{sec:ablations}

\paragraph{Cross-Scale Results (\cref{tab:ablations_4b}).}
On 4B, \ours achieves the highest OOD micro accuracy among the compared methods ($73.42\%$), exceeding GRPO by $0.97$ pp. This aggregate improvement does not extend to every benchmark: MMStar and Geo3K remain below Base.

\paragraph{Component Ablations (\cref{tab:ablations_4b}, bottom).}
All four variants reduce OOD micro accuracy below Base. Unpaired PCA improves in-domain accuracy but weakens OOD transfer, supporting the paired contrastive basis. A random basis tests the value of visual alignment, while the hard-subspace variant supports releasing the constraint after warmup. Single-view scoring removes both the second view and the consistency penalty, so its degradation measures their joint contribution.

\begin{table}[htbp]
  \centering
  \caption{\textbf{Subspace spectral probe and multi-seed Geometry3K diagnostics} on Qwen3-VL-4B ($r=32$). $|D|$: single-batch reward directional derivative; \textbf{Acc.}: test accuracy ($\%$, mean $\pm$ std).}
  \label{tab:probe}
  \small
  \setlength{\tabcolsep}{7.5pt}
  \renewcommand{\arraystretch}{1.14}
  \begin{adjustbox}{max width=\textwidth}
  \begin{tabular}{l ccccc}
    \toprule
    \textbf{Search Basis / View} & $C_{\mathrm{s}}\% \uparrow$ & $C_{\mathrm{n}}\% \downarrow$ & $\mathrm{SNR}\uparrow$ & $|D|\uparrow$ & \textbf{Geo3K Acc.} \\
    \midrule
    Isotropic Full-Param & 1.25 & 1.25 & 1.00 & -- & $64.7 \pm 0.6$ \\
    Ratio GEVD (Raw) & 0.12 & 0.01 & -- & 0.001 & \emph{Nullspace} \\
    Ratio GEVD (Reg.) & 53.3 & 36.1 & 1.48 & 0.053 & -- \\
    Unpaired PCA & \textbf{73.5} & 63.8 & 1.13 & 0.068 & $65.4 \pm 1.3$ \\
    Mild Scale ($s=0.96$) & -- & -- & -- & -- & $65.0 \pm 0.5$ \\
    Dual-View Iso. ($\lambda=0.1$) & 1.25 & 1.25 & 1.00 & -- & $66.6 \pm 0.6$ \\
    \rowcolor{oursrow}
    \textbf{Contrastive ($s=0.75$)} & 70.2 & \textbf{44.0} & \textbf{1.56} & \textbf{0.094} & \textbf{67.2 $\pm$ 0.4} \\
    \bottomrule
  \end{tabular}
  \end{adjustbox}
\end{table}

\subsection{Mechanistic Dissection}
\label{sec:mechanistic}

\begin{figure}[t]
  \centering
  \includegraphics[width=0.96\textwidth]{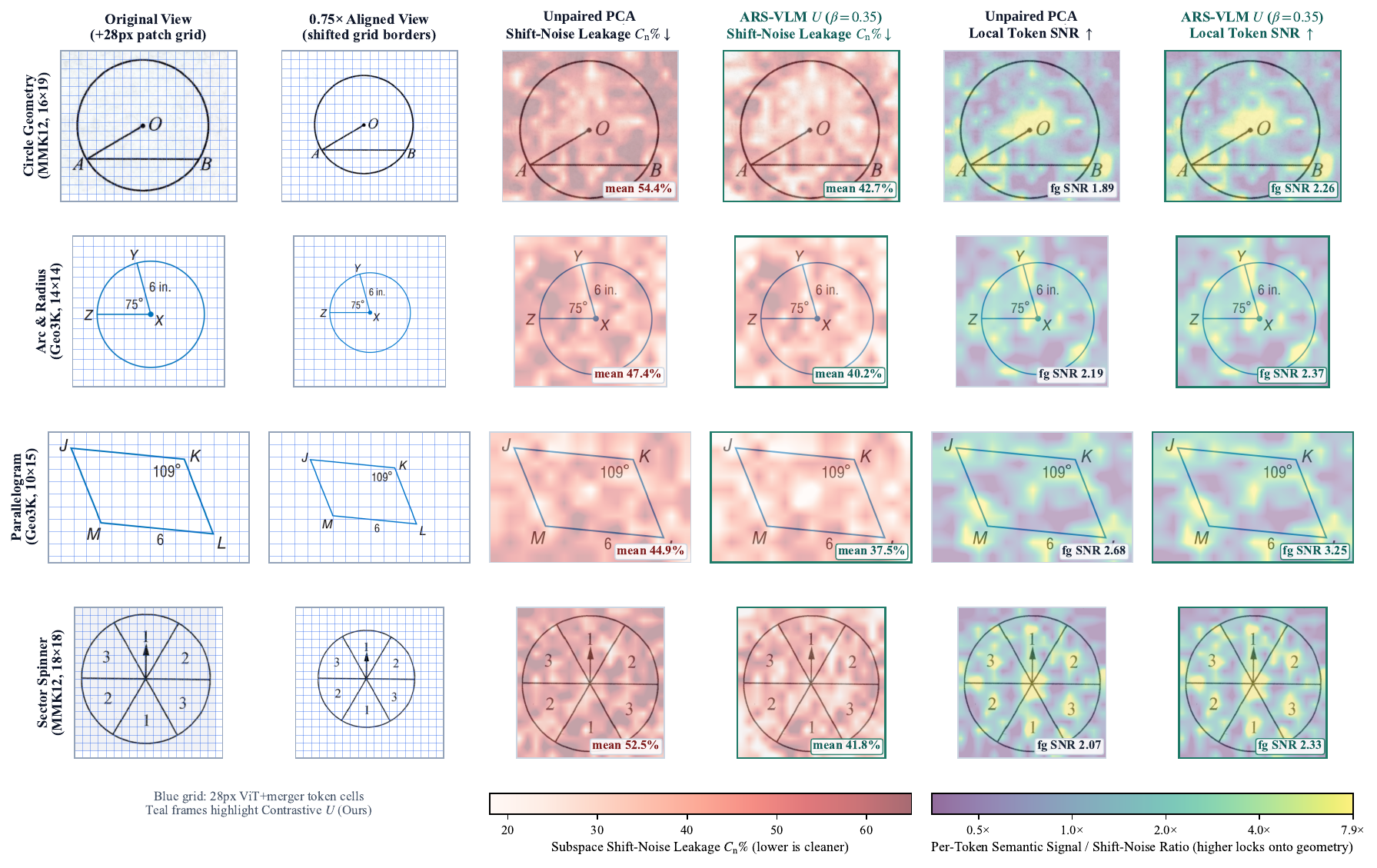}
  \caption{\textbf{Spatial visualization of visual alignment on Qwen3-VL-8B ($r=32$).} Columns show original and aligned views, followed by shift-noise leakage and local token SNR for Unpaired PCA and \ours ($\beta=0.35$). Across the four examples, \ours reduces mean leakage by $7.2$--$11.7$ percentage points and increases foreground SNR.}
  \label{fig:qualitative}
\end{figure}

\paragraph{Subspace SNR and Cross-Grid View Scale (\cref{tab:probe}).}
Verifying Proposition~\ref{prop:contrastive}, raw GEVD collapses into the nullspace ($0.12\%$ signal) and regularized GEVD loses $27\%$ signal ($|D|=0.053$), whereas our contrastive basis ($s=0.75$) retains $95.4\%$ of PCA's signal ($70.2\%$ vs.\ $73.5\%$) while cutting shift noise ($63.8\%\to 44.0\%$), boosting $|D|$ by $+37\%$ ($0.094$ vs.\ $0.068$), raising cross-view sign agreement to $87.5\%$, and halving Geometry3K seed variance ($67.2 \pm 0.4\%$).
Mild $0.96\times$ scaling leaves $66.6\%$ of tokens unshifted ($65.0\%$), whereas a $10$-scale sweep across $s\in[0.40,0.96]$ (Appendix~\ref{sec:supp_scale}) shows a broad invariance plateau across $s\in[0.65,0.85]$ ($92.1\%$--$100.0\%$ overlap with $U(0.75)$ on both 4B and 8B, vs.\ aliasing surge $\Tr(C_{\mathrm{n}})/\Tr(C_{\mathrm{s}})\to 1.90$ below $s=0.50$).

\paragraph{Qualitative Visual Alignment.}
Figure~\ref{fig:qualitative} shows four diagrams with their original and rescaled token grids, alongside spatial noise and SNR maps. Compared with Unpaired PCA, our contrastive basis reduces mean shift-noise leakage and increases foreground SNR in all four examples, illustrating how the learned subspace preserves visual signal under changes in patch-grid alignment.

\begin{figure}[htbp]
  \centering
  \includegraphics[width=0.72\textwidth]{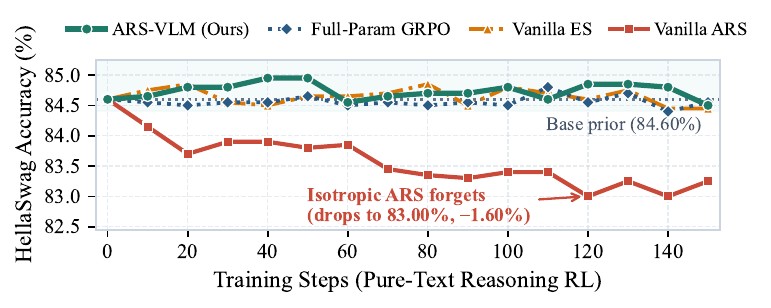}
  \caption{\textbf{Language prior retention during pure-text reasoning RL on Qwen3-VL-4B ($150$ steps).} Four-task mean HellaSwag accuracy across Countdown, GSM8K, MATH-500, and OlympiadBench. Isotropic ARS drops from $84.60\%$ to a minimum of $83.00\%$, whereas \ours stays close to the $84.60\%$ base prior (approximately $84.50\%$ at the final step).}
  \label{fig:forgetting}
\end{figure}

\paragraph{Static Memory, Warmup, and Policy KL (\cref{fig:dynamics}).}
On Qwen3-VL-8B, GRPO and \ours use approximately $64$ and $16.5\text{ GB/GPU}$ of static memory, respectively ($3.9\times$ ratio). Both figures exclude runtime activations, KV caches, and temporary buffers; component accounting and worker implementation details are in Appendix~\ref{sec:supp_memory}.
\ours reaches $61.0\%$ MMK12-Dev accuracy in $20$ steps versus GRPO's $150$, measuring optimization steps rather than wall-clock or query efficiency. Its shallow-layer weight drift is $31.1\%$ lower than vanilla ARS, and its policy KL estimate~\cite{schulman2020kl} changes only from $0.0392$ to $0.0398$ between steps $400$ and $500$. ARS reaches $0.0743$ at step $500$, while ES reaches $0.0620$ at step $1{,}000$.

\subsection{Cross-Modal and Linguistic Forgetting Suite}
\label{sec:forgetting}

\begin{table}[htbp]
  \centering
  \caption{\textbf{16-run pure-text reasoning and prior-forgetting suite on Qwen3-VL-4B} ($150$ steps/run). Peak target accuracy ($\%$) alongside worst-case retention ($\min$ across $15$ checkpoints) on HellaSwag (Base: $84.60\%$) and MMStar (Base: $57.67\%$); {\color{dropred}red} marks severe drops $\ge 1.5\%$ on HellaSwag ($<83.1\%$) or $\ge 3.0\%$ on MMStar ($<55.0\%$).}
  \label{tab:reasoning_4b}
  \small
  \setlength{\tabcolsep}{7.5pt}
  \renewcommand{\arraystretch}{1.12}
  \begin{adjustbox}{max width=\textwidth}
  \begin{tabular}{ll ccc c}
    \toprule
    \textbf{Task} (Base) & \textbf{Metric} & \textbf{GRPO} & \textbf{ES} & \textbf{ARS} & \ours \\
    \midrule
    \multirow{3}{*}{\makecell[l]{Countdown\\(66.60)}} & Target $\uparrow$ & 68.40 & \textbf{72.80} & 71.80 & 68.80 \\
    & HellaSwag$_{\min}$ & 84.40 & 84.00 & \color{dropred}82.80 & \textbf{84.60} \\
    & MMStar$_{\min}$ & \textbf{57.33} & \color{dropred}54.33 & \color{dropred}54.67 & 55.67 \\
    \midrule
    \multirow{3}{*}{\makecell[l]{GSM8K\\(90.20)}} & Target $\uparrow$ & 91.20 & \textbf{92.20} & 92.00 & 90.60 \\
    & HellaSwag$_{\min}$ & \textbf{84.20} & \textbf{84.20} & \color{dropred}83.00 & \textbf{84.20} \\
    & MMStar$_{\min}$ & 56.00 & \textbf{56.67} & \color{dropred}52.33 & 55.00 \\
    \midrule
    \multirow{3}{*}{\makecell[l]{MATH-500\\(65.20)}} & Target $\uparrow$ & 67.20 & \textbf{72.80} & 72.60 & 69.60 \\
    & HellaSwag$_{\min}$ & \textbf{84.40} & 83.60 & \color{dropred}81.80 & 84.20 \\
    & MMStar$_{\min / \mathrm{pk}}$ & 55.33 & 56.33 & 56.00 & \textbf{56.67 / 61.33} \\
    \midrule
    \multirow{3}{*}{\makecell[l]{Olympiad\\(21.67)}} & Target $\uparrow$ & 22.33 & \textbf{24.67} & 24.33 & 23.00 \\
    & HellaSwag$_{\min}$ & \textbf{84.20} & \textbf{84.20} & \color{dropred}82.60 & \textbf{84.20} \\
    & MMStar$_{\min / \mathrm{end}}$ & 56.33 & 55.67 & \color{dropred}53.67 & \textbf{57.00 / 59.33} \\
    \bottomrule
  \end{tabular}
  \end{adjustbox}
\end{table}

\cref{tab:reasoning_4b} and \cref{fig:forgetting} evaluate linguistic and visual retention across $16$ pure-text RL runs ($600$ steps/method).
Isotropic ARS/ES reach higher pure-text peaks, but ARS loses $2.8$ pp on HellaSwag ($84.6\%\to81.8\%$ on MATH) and $5.34$ pp on zero-shot MMStar ($57.67\%\to52.33\%$ on GSM8K).
Anchoring shallow visual-fusion layers ($\rho_{\mathrm{soft}}=0.25$) allows \ours to beat GRPO on \textbf{three of four} target reasoning benchmarks ($+2.4$ pp on MATH-500), with at most $0.4$ pp HellaSwag loss in these runs. Visual retention varies by task: MMStar reaches $61.33\%$ at the MATH peak and $59.33\%$ at the OlympiadBench endpoint, but its GSM8K minimum is $55.00\%$, below the $57.67\%$ base.

\section{Conclusion and Limitations}
\label{sec:conclusion}

\ours combines a contrastive visual basis, a layer-selective soft prior, and dual-view selection for forward-only VLM post-training. The visual basis suppresses patch-grid shift noise, while the soft prior preserves flexibility in deeper reasoning layers. On Qwen3-VL-4B and 8B, this design improves OOD retention; the validation-selected 8B checkpoint reaches $75.41\%$ single-view OOD micro accuracy, comparable to GRPO at lower static memory cost. These results support using cross-view visual structure to guide parameter-space search without backpropagation or a reference policy.

\paragraph{Limitations.}
Our evaluation covers two sizes of one model family, primarily on multimodal STEM tasks; transfer to other architectures and broader visual domains remains untested. The visual prior reduces peak accuracy on some text tasks, and dual-view scoring increases rollout cost. Faster warmup in optimization steps does not establish a wall-clock or reward-query advantage. Future work should compare equal-query budgets and adapt the subspace rank and prior strength to each layer and task.

\bibliographystyle{ieeenat_fullname}
\bibliography{main}

@inproceedings{mania2018ars,
  author    = {Horia Mania and Aurelia Guy and Benjamin Recht},
  title     = {Simple Random Search of Static Linear Policies is Competitive for Reinforcement Learning},
  booktitle = {Advances in Neural Information Processing Systems (NeurIPS)},
  volume    = {31},
  year      = {2018}
}

@inproceedings{malladi2023mezo,
  author    = {Sadhika Malladi and Tianyu Gao and Eshaan Nichani and Alex Damian and Jason D. Lee and Danqi Chen and Sanjeev Arora},
  title     = {Fine-Tuning Language Models with Just Forward Passes},
  booktitle = {Advances in Neural Information Processing Systems (NeurIPS)},
  volume    = {36},
  year      = {2023}
}

@inproceedings{maheswaranathan2019guided,
  author    = {Niru Maheswaranathan and Luke Metz and George Tucker and Dami Choi and Jascha Sohl-Dickstein},
  title     = {Guided Evolutionary Strategies: Augmenting Random Search with Surrogate Gradients},
  booktitle = {International Conference on Machine Learning (ICML)},
  pages     = {4264--4273},
  year      = {2019}
}

@article{spall1992spsa,
  author  = {James C. Spall},
  title   = {Multivariate Stochastic Approximation Using a Simultaneous Perturbation Gradient Approximation},
  journal = {IEEE Transactions on Automatic Control},
  volume  = {37},
  number  = {3},
  pages   = {332--341},
  year    = {1992}
}

@article{salimans2017es,
  author  = {Tim Salimans and Jonathan Ho and Xi Chen and Szymon Sidor and Ilya Sutskever},
  title   = {Evolution Strategies as a Scalable Alternative to Reinforcement Learning},
  journal = {arXiv preprint arXiv:1703.03864},
  year    = {2017}
}

@article{nesterov2017random,
  author  = {Yurii Nesterov and Vladimir Spokoiny},
  title   = {Random Gradient-Free Minimization of Convex Functions},
  journal = {Foundations of Computational Mathematics},
  volume  = {17},
  number  = {2},
  pages   = {527--566},
  year    = {2017}
}

@article{qiu2025esscale,
  author  = {Xin Qiu and Yulu Gan and Conor F. Hayes and Qiyao Liang and Elliot Meyerson and Babak Hodjat and Risto Miikkulainen},
  title   = {Evolution Strategies at Scale: {LLM} Fine-Tuning Beyond Reinforcement Learning},
  journal = {arXiv preprint arXiv:2509.24372},
  year    = {2025}
}

@article{sun2026essam,
  author  = {Zhishen Sun and Sizhe Dang and Guang Dai and Haishan Ye},
  title   = {{ESSAM}: A Novel Competitive Evolution Strategies Approach to Reinforcement Learning for Memory Efficient {LLMs} Fine-Tuning},
  journal = {arXiv preprint arXiv:2602.01003},
  year    = {2026}
}

@inproceedings{abdi2026forgetting,
  author    = {Immanuel Abdi and Akshat Gupta and Micah Mok and Alexander Lu and Nicholas Lee and Gopala Anumanchipalli},
  title     = {Evolutionary Strategies at Scale Lead to Catastrophic Forgetting},
  booktitle = {Proceedings of the Annual Meeting of the Association for Computational Linguistics (ACL)},
  note      = {arXiv:2601.20861},
  year      = {2026}
}

@article{agzo2026,
  author  = {Yifan Zhang and others},
  title   = {Activation-Guided Zeroth-Order Optimization for Large Language Model Fine-Tuning},
  journal = {arXiv preprint arXiv:2601.17261},
  year    = {2026}
}

@article{shao2024deepseekmath,
  author  = {Zhihong Shao and Peiyi Wang and Qihao Zhu and Runxin Xu and Junxiao Song and Xiao Bi and Haowei Zhang and Mingchuan Zhang and Y. K. Li and Y. Wu and Daya Guo},
  title   = {{DeepSeekMath}: Pushing the Limits of Mathematical Reasoning in Open Language Models},
  journal = {arXiv preprint arXiv:2402.03300},
  year    = {2024}
}

@article{guo2025deepseekr1,
  author  = {Daya Guo and Dejian Yang and Haowei Zhang and Junxiao Song and Ruoyu Zhang and Runxin Xu and Qihao Zhu and Shirong Ma and Peiyi Wang and Xiao Bi and others},
  title   = {{DeepSeek-R1}: Incentivizing Reasoning Capability in {LLMs} via Reinforcement Learning},
  journal = {arXiv preprint arXiv:2501.12948},
  year    = {2025}
}

@article{consistroll2026,
  author  = {Zhe Chen and others},
  title   = {{ConsistRoll}: View-Consistent Policy Rollouts for Multimodal Reasoning},
  journal = {arXiv preprint arXiv:2606.29812},
  year    = {2026}
}

@article{sage2026,
  author  = {Haoran Wang and others},
  title   = {{SAGE}: Semantically Aligned Group Policy Optimization for Vision-Language Reasoning},
  journal = {arXiv preprint arXiv:2605.18162},
  year    = {2026}
}

@article{hu2022lora,
  author  = {Edward J. Hu and Yelong Shen and Phillip Wallis and Zeyuan Allen-Zhu and Yuanzhi Li and Shean Wang and Lu Wang and Weizhu Chen},
  title   = {{LoRA}: Low-Rank Adaptation of Large Language Models},
  journal = {International Conference on Learning Representations (ICLR)},
  year    = {2022}
}

@article{schulman2025lorawithoutregret,
  author  = {John Schulman and Thinking Machines Lab},
  title   = {{LoRA} Without Regret},
  journal = {Thinking Machines Lab Technical Blog},
  year    = {2025}
}

@misc{schulman2020kl,
  author       = {John Schulman},
  title        = {Approximating {KL} Divergence},
  howpublished = {\url{http://joschu.net/blog/kl-approx.html}},
  year         = {2020}
}

@inproceedings{abid2018contrastive,
  author    = {Abubakar Abid and Martin J. Zhang and Vivek K. Bagaria and James Zou},
  title     = {Exploring Patterns Enriched in a Dataset with Contrastive Principal Component Analysis},
  booktitle = {Nature Communications},
  volume    = {9},
  pages     = {2134},
  year      = {2018}
}

@article{qwen3vl2025,
  author  = {Qwen Team},
  title   = {{Qwen3-VL} Technical Report},
  journal = {arXiv preprint},
  year    = {2025}
}

@inproceedings{kwon2023vllm,
  author    = {Woosuk Kwon and Zhuohan Li and Siyuan Zhuang and Ying Sheng and Lianmin Zheng and Cody Hao Yu and Joseph E. Gonzalez and Hao Zhang and Ion Stoica},
  title     = {Efficient Memory Management for Large Language Model Serving with {PagedAttention}},
  booktitle = {Symposium on Operating Systems Principles (SOSP)},
  year      = {2023}
}

@article{meng2025mmeureka,
  author  = {Fanqing Meng and Lingxiao Du and Zongkai Liu and Zhixiang Zhou and Quanfeng Lu and Daocheng Tu and Tiaoyang Ye and Qi Wang and Yu Qiao and Ping Luo and Kaipeng Zhang},
  title   = {{MM-Eureka}: Exploring the Frontiers of Multimodal Reasoning with Rule-based Reinforcement Learning},
  journal = {arXiv preprint arXiv:2503.07365},
  year    = {2025}
}

@inproceedings{lu2021intergps,
  author    = {Pan Lu and Ran Gong and Shibiao Jiang and Liang Qiu and Siyuan Huang and Xiaodan Liang and Song-Chun Zhu},
  title     = {{Inter-GPS}: Interpretable Geometry Problem Solving with Formal Language and Symbolic Reasoning},
  booktitle = {Proceedings of the Joint Conference of the Annual Meeting of the Association for Computational Linguistics (ACL-IJCNLP)},
  year      = {2021}
}

@inproceedings{lu2024mathvista,
  author    = {Pan Lu and Hritik Bansal and Tony Xia and Jiacheng Liu and Chunyuan Li and Hannaneh Hajishirzi and Hao Cheng and Kai-Wei Chang and Michel Galley and Jianfeng Gao},
  title     = {{MathVista}: Evaluating Mathematical Reasoning of Foundation Models in Visual Contexts},
  booktitle = {International Conference on Learning Representations (ICLR)},
  year      = {2024}
}

@inproceedings{qiao2024wemath,
  author    = {Runqi Qiao and Qiuna Tan and Guanting Dong and Minhui Wu and Chong Sun and Xiaoshuai Song and Zhuoma GongQue and Shanglin Lei and Zhe Wei and Miaoxuan Zhang and others},
  title     = {{We-Math}: Does Your Large Multimodal Model Achieve Human-like Mathematical Reasoning?},
  booktitle = {Annual Meeting of the Association for Computational Linguistics (ACL)},
  year      = {2025}
}

@inproceedings{lu2022scienceqa,
  author    = {Pan Lu and Swaroop Mishra and Tanglin Xia and Liang Qiu and Kai-Wei Chang and Song-Chun Zhu and Oyvind Tafjord and Peter Clark and Ashwin Kalyan},
  title     = {Learn to Explain: Multimodal Reasoning via Thought Chains for Science Question Answering},
  booktitle = {Advances in Neural Information Processing Systems (NeurIPS)},
  volume    = {35},
  year      = {2022}
}

@inproceedings{chen2024mmstar,
  author    = {Lin Chen and Jinsong Li and Xiaoyi Dong and Pan Zhang and Yuhang Zang and Zehui Chen and Haodong Duan and Jiaqi Wang and Yu Qiao and Dahua Lin and Feng Zhao},
  title     = {Are We on the Right Way for Evaluating Large Vision-Language Models?},
  booktitle = {Advances in Neural Information Processing Systems (NeurIPS)},
  volume    = {37},
  year      = {2024}
}

@inproceedings{wang2024mathvision,
  author    = {Ke Wang and Junting Pan and Weikang Shi and Zimu Lu and Mingjie Zhan and Hongsheng Li},
  title     = {Measuring Multimodal Mathematical Reasoning with {MATH-Vision} Dataset},
  booktitle = {Advances in Neural Information Processing Systems (NeurIPS)},
  volume    = {37},
  year      = {2024}
}

@article{xiao2024logicvista,
  author  = {Yijia Xiao and Edward Sun and Tianyu Liu and Wei Wang},
  title   = {{LogicVista}: Multimodal {LLM} Logical Reasoning Benchmark in Visual Contexts},
  journal = {arXiv preprint arXiv:2407.04973},
  year    = {2024}
}

@inproceedings{he2024olympiadbench,
  author    = {Chaoqun He and Renjie Luo and Yuzhuo Bai and Shengding Hu and Zhen Leng Thai and Junhao Shen and Jinyi Hu and Xu Han and Yujie Huang and Yuxiang Zhang and others},
  title     = {{OlympiadBench}: A Challenging Benchmark for Promoting {AGI} with Olympiad-Level Bilingual Multimodal Scientific Problems},
  booktitle = {Annual Meeting of the Association for Computational Linguistics (ACL)},
  year      = {2024}
}

@article{cobbe2021gsm8k,
  author  = {Karl Cobbe and Vineet Kosaraju and Mohammad Bavarian and Mark Chen and Heewoo Jun and Lukasz Kaiser and Matthias Plappert and Jerry Tworek and Jacob Hilton and Reiichiro Nakano and others},
  title   = {Training Verifiers to Solve Math Word Problems},
  journal = {arXiv preprint arXiv:2110.14168},
  year    = {2021}
}

@inproceedings{hendrycks2021math,
  author    = {Dan Hendrycks and Collin Burns and Saurav Kadavath and Akul Arora and Steven Basart and Eric Tang and Dawn Song and Jacob Steinhardt},
  title     = {Measuring Mathematical Problem Solving With the {MATH} Dataset},
  booktitle = {Advances in Neural Information Processing Systems Track on Datasets and Benchmarks},
  year      = {2021}
}

@inproceedings{zellers2019hellaswag,
  author    = {Rowan Zellers and Ari Holtzman and Yonatan Bisk and Ali Farhadi and Yejin Choi},
  title     = {{HellaSwag}: Can a Machine Really Finish Your Sentence?},
  booktitle = {Annual Meeting of the Association for Computational Linguistics (ACL)},
  year      = {2019}
}

@inproceedings{zhang2024zo_subspace,
  author    = {Yihua Zhang and Pingzhi Li and Junyuan Hong and Jiaxiang Li and Yimeng Zhang and Wenqing Zheng and Pin-Yu Chen and Jason D. Lee and Wotao Yin and Mingyi Hong and Zhangyang Wang and Sijia Liu},
  title     = {Revisiting Zeroth-Order Optimization for Memory-Efficient {LLM} Fine-Tuning: A Benchmark},
  booktitle = {International Conference on Machine Learning (ICML)},
  year      = {2024}
}

@article{wang2026prism,
  author  = {Sudong Wang and Weiquan Huang and Xiaomin Yu and Zuhao Yang and Hehai Lin and Keming Wu and Chaojun Xiao and Chen Chen and Wenxuan Wang and Beier Zhu and Yunjian Zhang and Chengwei Qin},
  title   = {Beyond {SFT}-to-{RL}: Pre-alignment via Black-Box On-Policy Distillation for Multimodal {RL}},
  journal = {arXiv preprint arXiv:2604.28123},
  year    = {2026}
}

\makeappendixbanner

\section{Proof of Proposition~\ref{prop:contrastive}}
\label{sec:supp_proof}

\begin{proof}
\textbf{Part 1 (Cross-covariance decomposition).}
Let $(f_{\mathrm{o}}, \tilde{f}_{\mathrm{a}}) \in \mathbb{R}^d \times \mathbb{R}^d$ be a zero-mean bilinearly paired token pair across the original and $0.75\times$ aligned views, with view-average $s = \frac{1}{2}(f_{\mathrm{o}} + \tilde{f}_{\mathrm{a}})$ and view-shift residual $n = f_{\mathrm{o}} - \tilde{f}_{\mathrm{a}}$.
Recall that $C_{\mathrm{pool}} = \frac{1}{2}\mathbb{E}[f_{\mathrm{o}} f_{\mathrm{o}}^\top + \tilde{f}_{\mathrm{a}}\tilde{f}_{\mathrm{a}}^\top]$ and $C_{\mathrm{cross}} = \frac{1}{2}\mathbb{E}[f_{\mathrm{o}}\tilde{f}_{\mathrm{a}}^\top + \tilde{f}_{\mathrm{a}} f_{\mathrm{o}}^\top]$.
Expanding the second moments gives
\begin{align}
    C_{\mathrm{s}} &= \mathbb{E}[s s^\top] = \frac{1}{4}\mathbb{E}\big[f_{\mathrm{o}} f_{\mathrm{o}}^\top + \tilde{f}_{\mathrm{a}}\tilde{f}_{\mathrm{a}}^\top + f_{\mathrm{o}}\tilde{f}_{\mathrm{a}}^\top + \tilde{f}_{\mathrm{a}} f_{\mathrm{o}}^\top\big] \nonumber\\
    &= \frac{1}{2} C_{\mathrm{pool}} + \frac{1}{2} C_{\mathrm{cross}}, \\
    C_{\mathrm{n}} &= \mathbb{E}[n n^\top] = \mathbb{E}\big[f_{\mathrm{o}} f_{\mathrm{o}}^\top + \tilde{f}_{\mathrm{a}}\tilde{f}_{\mathrm{a}}^\top - f_{\mathrm{o}}\tilde{f}_{\mathrm{a}}^\top - \tilde{f}_{\mathrm{a}} f_{\mathrm{o}}^\top\big] \nonumber\\
    &= 2 C_{\mathrm{pool}} - 2 C_{\mathrm{cross}}.
\end{align}
Substituting both expressions into $M_\beta = C_{\mathrm{s}} - \beta C_{\mathrm{n}}$ yields
\begin{equation}
    M_\beta = \left(\frac{1}{2} - 2\beta\right) C_{\mathrm{pool}} + \left(\frac{1}{2} + 2\beta\right) C_{\mathrm{cross}},
\end{equation}
which proves \cref{eq:cross_cov_id}.
Setting $\beta = 0$ gives $M_0 = C_{\mathrm{s}} = \frac{1}{2}(C_{\mathrm{pool}} + C_{\mathrm{cross}})$, the covariance of paired cross-view means, whereas the Unpaired PCA baseline in the main text uses the top eigenvectors of $C_{\mathrm{pool}}$ directly.
At $\beta = \frac{1}{4}$, $M_{1/4} = C_{\mathrm{cross}}$ isolates the symmetric cross-view covariance. For $\beta > \frac{1}{4}$ (such as our default $\beta = 0.35$, where $M_{0.35} = 1.2\,C_{\mathrm{cross}} - 0.2\,C_{\mathrm{pool}}$), the negative weight on $C_{\mathrm{pool}}$ penalizes view-specific variance without inverting $(C_{\mathrm{n}} + \epsilon I)$.

\medskip
\noindent\textbf{Part 2 (Frobenius norm conservation and directional gain).}
Let $U \in \mathbb{R}^{d\times r}$ satisfy $U^\top U = I_r$, and let $Z \in \mathbb{R}^{d_{\mathrm{out}}\times r}$ have i.i.d.\ $\mathcal{N}(0, 1)$ entries.
For $\Delta W = \sigma_{\mathrm{noise}}\sqrt{d/r}\, Z U^\top$, cyclic trace invariance gives
\begin{align}
    \mathbb{E}\|\Delta W\|_F^2 &= \sigma_{\mathrm{noise}}^2 \frac{d}{r}\, \mathbb{E}\!\left[\Tr(U Z^\top Z U^\top)\right] \nonumber\\
    &= \sigma_{\mathrm{noise}}^2 \frac{d}{r}\, \Tr\!\big(U^\top U\, \mathbb{E}[Z^\top Z]\big) \nonumber\\
    &= \sigma_{\mathrm{noise}}^2 \frac{d}{r}\, \Tr(I_r \cdot d_{\mathrm{out}} I_r) = \sigma_{\mathrm{noise}}^2 d_{\mathrm{out}} d,
\end{align}
matching the expected squared Frobenius norm of an isotropic perturbation $\Delta W_{\mathrm{iso}} \sim \mathcal{N}(0, \sigma_{\mathrm{noise}}^2 I_{d_{\mathrm{out}}\times d})$.
For the Stage~2 mixture $\Delta W_\rho = \sigma_{\mathrm{noise}}(\sqrt{1-\rho^2} Z_{\mathrm{full}} + \rho\sqrt{d/r} Z_{\mathrm{sub}} U^\top)$ with independent $Z_{\mathrm{full}}$ and $Z_{\mathrm{sub}}$, the cross term has zero expectation and
\begin{align}
    \mathbb{E}\|\Delta W_\rho\|_F^2 &= \sigma_{\mathrm{noise}}^2 \big((1-\rho^2) d_{\mathrm{out}} d + \rho^2 d_{\mathrm{out}} d\big) \nonumber\\
    &= \sigma_{\mathrm{noise}}^2 d_{\mathrm{out}} d.
\end{align}

Finally, decompose $G = \nabla_W \mathcal{J} \in \mathbb{R}^{d_{\mathrm{out}}\times d}$ into $G = G_\parallel + G_\perp$, where $G_\parallel = G U U^\top$ lies in $\operatorname{span}(U)$ and $G_\perp U = 0$.
Under isotropic perturbations, $\mathbb{E}[\langle G_\parallel, \Delta W_{\mathrm{iso}}\rangle^2] = \sigma_{\mathrm{noise}}^2 \|G_\parallel\|_F^2$.
Under $\Delta W = \sigma_{\mathrm{noise}}\sqrt{d/r}\, Z U^\top$, we have $\langle G, \Delta W\rangle = \sigma_{\mathrm{noise}}\sqrt{d/r}\,\langle G U, Z\rangle$, and therefore
\begin{equation}
    \mathbb{E}\big[\langle G, \Delta W\rangle^2\big] = \sigma_{\mathrm{noise}}^2 \frac{d}{r} \|G U\|_F^2 = \frac{d}{r}\, \sigma_{\mathrm{noise}}^2 \|G_\parallel\|_F^2.
\end{equation}
For the Stage~2 mixture $\Delta W_\rho$, any unit-norm matrix $V \in \mathbb{R}^{d_{\mathrm{out}}\times d}$ supported in $\operatorname{span}(U)$ ($V = V U U^\top$, $\|V\|_F = 1$) satisfies $\mathbb{E}[\langle V, \Delta W_\rho\rangle^2] = \sigma_{\mathrm{noise}}^2 (1 - \rho^2 + \rho^2 \frac{d}{r})$. At $\rho_{\mathrm{soft}} = 0.25$ and $r = 32$, the subspace term contributes $\rho_{\mathrm{soft}}^2 d/r = 5\times$ on 4B ($d=2560$) and $8\times$ on 8B ($d=4096$), yielding total directional variance factors of $5.9375\times$ and $8.9375\times$, respectively.
\end{proof}

\section{Static Memory Accounting and Worker Implementation}
\label{sec:supp_memory}

\begin{table}[htbp]
  \centering
  \caption{\textbf{Static memory accounting on Qwen3-VL-8B-Instruct} ($8.77\text{B}$ total parameters; $0.42\text{B}$ frozen vision patch/block weights and $8.35\text{B}$ trainable merger and decoder weights). Sizes are reported in binary gigabytes ($2^{30}$ bytes), matching the rounded ${\sim}64$ and ${\sim}16.5\text{ GB/GPU}$ figures in the main text, and exclude runtime activations, KV caches, and temporary execution buffers.}
  \label{tab:supp_memory}
  \small
  \setlength{\tabcolsep}{9.0pt}
  \renewcommand{\arraystretch}{1.12}
  \begin{adjustbox}{max width=\textwidth}
  \begin{tabular}{l cc}
    \toprule
    \textbf{Allocated Static Component} & \textbf{GRPO Trainer GPU} & \textbf{\ours Worker GPU} \\
    \midrule
    Policy weights (\texttt{bf16}, $8.77\text{B}$) & $16.3\text{ GB}$ & $16.3\text{ GB}$ \\
    Reference policy $\pi_{\mathrm{ref}}$ (\texttt{bf16}, shared ViT) & $15.6\text{ GB}$ & $0$ \\
    Sharded gradients ($8.35\text{B} / 4$, \texttt{fp32}) & $7.8\text{ GB}$ & $0$ \\
    ZeRO-1 \texttt{fp32} master weights ($8.35\text{B} / 4$) & $7.8\text{ GB}$ & $0$ \\
    ZeRO-1 \texttt{fp32} AdamW moments ($m, v$) & $15.6\text{ GB}$ & $0$ \\
    Contrastive visual basis $U$ ($4096 \times 32$) & $0$ & $<0.5\text{ MB}$ \\
    \midrule
    \textbf{Total static footprint (incl.\ runtime metadata)} & $63.1\text{ GB}$ (${\sim}64\text{ GB}$) & $16.3\text{ GB}$ (${\sim}16.5\text{ GB}$) \\
    \bottomrule
  \end{tabular}
  \end{adjustbox}
\end{table}

\cref{tab:supp_memory} provides the component breakdown behind the static memory comparison in \cref{fig:dynamics}a and \cref{sec:mechanistic}.
Qwen3-VL-8B-Instruct contains $8.77\text{B}$ parameters ($0.42\text{B}$ in the vision patch embedder and transformer blocks, $0.16\text{B}$ in the spatial mergers, and $8.19\text{B}$ in the language decoder and embeddings).

All training and evaluation experiments in this work are conducted on $8 \times \text{NVIDIA A100}$ ($80\text{ GB}$) GPUs.
In our full-parameter GRPO setup on $8 \times \text{A100}$, four GPUs host \texttt{vLLM} rollout engines ($G=8$ trajectories per problem) and four GPUs run distributed PyTorch training with activation checkpointing and ZeRO-1 sharded AdamW.
Following standard VLM RLVR practice, the vision patch/block backbone ($0.42\text{B}$) is frozen and shared in memory between the policy and $\pi_{\mathrm{ref}}$, leaving $8.35\text{B}$ trainable parameters.
Each trainer GPU holds a \texttt{bf16} copy of the policy ($16.3\text{ GB}$), the non-shared \texttt{bf16} weights of $\pi_{\mathrm{ref}}$ ($15.6\text{ GB}$), and one-fourth of the \texttt{fp32} master weights, gradients, and first/second AdamW moments ($31.2\text{ GB}$ combined), totaling $63.1\text{ GB}$ (${\sim}64\text{ GB/GPU}$ with distributed state metadata) before counting activation checkpoints or unsharded backward gradient buffers, alongside per-step weight synchronization from the trainer to the \texttt{vLLM} workers.

By contrast, \ours and all forward-only ZO baselines run across all eight A100 GPUs as persistent \texttt{vLLM} inference workers without a separate trainer fleet. Each worker stores only the \texttt{bf16} model weights ($16.3\text{ GB}$, or ${\sim}16.5\text{ GB}$ with engine parameter metadata) and the basis matrix $U \in \mathbb{R}^{4096\times 32}$ ($0.25\text{ MB}$ in \texttt{bf16}, $0.50\text{ MB}$ in \texttt{fp32}).
During each step, workers apply $\theta_t \pm \Delta\theta_i$ in-place by regenerating Gaussian blocks from integer seeds, reset multimodal prefix caches so visual merger features reflect updated weights, decode paired views with PagedAttention, restore $\theta_t$, and accumulate the selected top-$k$ update tensor-by-tensor in \texttt{fp32} before casting back to \texttt{bf16}.
As noted in the main text, both static figures omit runtime forward activations, KV caches, and temporary workspace buffers, which depend on batch concurrency and sequence length.

\section{Dual-View Truncation Fallback at 4,096 Tokens}
\label{sec:supp_dv}

\begin{table}[t]
  \centering
  \caption{\textbf{Label-free Dual-View Truncation Fallback (DV-Fallback) at the $4{,}096$-token limit across Qwen3-VL-8B-Instruct post-training and public RLVR checkpoints.} \textbf{Trig.\ $\%$} is the fraction of OOD items lacking a closing \texttt{\textbackslash boxed\{\}} delimiter on the original view, which are re-evaluated once on the $0.75\times$ aligned view without using ground-truth labels. Following \cref{tab:main_8b}, first-order RL (\textbf{FO}; adding public \texttt{NoisyRollout-MMK12-6.4K-7B}) is listed as a reference and best \textbf{ZO} in \textbf{bold}; public RLVR rows in \cref{tab:main_8b} already reflect this DV-Fallback cache, whereas all our 8B runs there are strictly single-view. The 8B Unpaired PCA baseline from \cref{tab:main_8b} is omitted because paired-view rollout caches were not retained for that run.}
  \label{tab:supp_dv_fallback}
  \footnotesize
  \setlength{\tabcolsep}{4.0pt}
  \renewcommand{\arraystretch}{1.10}
  \begin{adjustbox}{max width=\textwidth}
  \begin{tabular}{l c c ccccccc cc c}
    \toprule
    \multirow{2}{*}{\textbf{Method (8B \& Public RLVR)}} & \multirow{2}{*}{\textbf{Mode}} & \textbf{Trig.} & \multicolumn{7}{c}{\textbf{DV-Fallback Accuracy ($\%$) Across Seven OOD Benchmarks}} & \multicolumn{2}{c}{\textbf{DV 7-OOD}} & \textbf{MMK12} \\
    \cmidrule(lr){4-10} \cmidrule(lr){11-12}
    & & $\%$ & MathVista & We-Math & SciQA & MMStar & Geo3K & MathVis & LogicVis & \textbf{Macro} & \textbf{Micro} & \textbf{Test} \\
    \midrule
    Pretrained Base (Step 0) & -- & 14.2 & 77.10 & 74.31 & 94.75 & 71.33 & 58.40 & 37.50 & 55.58 & 67.00 & 75.68 & 53.25 \\
    \midrule
    \multicolumn{13}{l}{\emph{First-Order RL: Public RLVR Checkpoints and Our Full-Parameter GRPO Baseline}} \\
    MM-Eureka-Qwen-7B~\cite{meng2025mmeureka} & FO & 11.2 & 69.10 & 63.33 & 89.14 & 63.80 & 40.77 & 32.24 & 44.87 & 57.61 & 66.91 & 57.65 \\
    NoisyRollout-MMK12-7B & FO & 31.8 & 70.60 & 70.29 & 89.29 & 65.33 & 45.42 & 27.63 & 41.29 & 58.55 & 69.01 & 54.05 \\
    PRISM-DAPO-8B~\cite{wang2026prism} & FO & 47.7 & 43.60 & 28.62 & 79.33 & 51.47 & 42.26 & 20.39 & 16.52 & 40.31 & 48.57 & 7.20 \\
    Full-Param GRPO (Best)~\cite{shao2024deepseekmath} & FO & 6.9 & 77.00 & 76.67 & 94.89 & 71.00 & 61.56 & 43.09 & 60.05 & 69.18 & 76.91 & 58.30 \\
    Full-Param GRPO (Step 500)~\cite{shao2024deepseekmath} & FO & 6.9 & 78.80 & 77.07 & 94.69 & 71.33 & 60.90 & 45.07 & 57.59 & 69.35 & 77.14 & 56.10 \\
    \midrule
    \multicolumn{13}{l}{\emph{Forward-Only Zeroth-Order Baselines \& Subspace Ablations (Zero Gradients, Zero Optimizer States, No $\pi_{\mathrm{ref}}$, ${\sim}16.5\text{ GB/GPU}$ Static)}} \\
    MeZO-RGE ($N=16$, Step 500)~\cite{malladi2023mezo} & ZO & 12.1 & 77.00 & 75.46 & 94.65 & 70.80 & 60.73 & 38.49 & 54.91 & 67.43 & 75.98 & 54.90 \\
    Vanilla ES (Best)~\cite{qiu2025esscale} & ZO & 9.6 & 74.80 & 75.98 & 93.90 & \textbf{71.00} & 59.90 & 41.45 & 54.24 & 67.32 & 75.66 & 51.75 \\
    Vanilla ES (Step 500)~\cite{qiu2025esscale} & ZO & 8.8 & 73.70 & 76.38 & 93.75 & 68.27 & 58.74 & 39.80 & 55.80 & 66.63 & 74.97 & 51.15 \\
    Vanilla ARS (Best, Step 90)~\cite{mania2018ars} & ZO & 10.6 & \textbf{77.20} & 76.03 & 94.75 & 70.40 & \textbf{61.07} & 41.12 & 55.80 & 68.05 & 76.27 & \textbf{58.65} \\
    Vanilla ARS (Step 500)~\cite{mania2018ars} & ZO & 8.9 & 67.70 & 71.95 & 88.99 & 64.33 & 53.41 & 34.54 & 53.35 & 62.04 & 70.35 & 53.55 \\
    Hard Subspace ($\rho=1.0$, Step 500) & ZO & 5.7 & 66.30 & 67.99 & 87.46 & 64.60 & 45.09 & 33.88 & 54.02 & 59.91 & 68.27 & 48.40 \\
    Random Orthonormal Subspace (Best) & ZO & 9.2 & 76.80 & 77.13 & 95.24 & 69.00 & 60.23 & 40.46 & 52.90 & 67.39 & 76.06 & 52.70 \\
    \rowcolor{oursrow}
    \ours \textbf{(Ours, Best)} & \textbf{ZO} & 7.1 & 76.80 & 77.59 & \textbf{95.59} & 70.67 & 58.24 & \textbf{43.75} & \textbf{58.48} & \textbf{68.73} & \textbf{76.89} & 56.65 \\
    \rowcolor{oursrow}
    \ours \textbf{(Ours, Step 500)} & \textbf{ZO} & 7.1 & 76.10 & \textbf{77.76} & 94.79 & 70.67 & 60.23 & 42.76 & 54.69 & 68.14 & 76.52 & 56.85 \\
    \bottomrule
  \end{tabular}
  \end{adjustbox}
\end{table}

When a diagram feature falls near a $28\text{px}$ patch-merger boundary, the model occasionally enters repetitive re-reading and reaches the $4{,}096$-token limit without producing a final \texttt{\textbackslash boxed\{\}} answer.
Because \texttt{\textbackslash boxed\{\}} presence is observable from the raw output string without ground-truth labels, DV-Fallback re-queries the model on the deterministic $0.75\times$ aligned view only when the original response contains no \texttt{\textbackslash boxed\{\}} delimiter.

\cref{tab:supp_dv_fallback} reports the resulting performance when this rule is applied uniformly across Qwen3-VL-8B checkpoints and public RLVR models at $4{,}096$ tokens.
For \ours (Best), only $7.1\%$ of OOD test items trigger the second view ($1.071\times$ average inference calls per problem, down from $14.2\%$ on Base, $11.2\%\text{--}47.7\%$ across public RLVR checkpoints, and $10.6\%$ on Vanilla ARS Best), raising 7-OOD micro accuracy from $75.41\%$ (\cref{tab:main_8b}) to $76.89\%$ and MMK12-Test accuracy from $53.70\%$ to $56.65\%$ ($56.85\%$ at step $500$).
Full-Param GRPO (Best) triggers fallback on $6.9\%$ of items and reaches $76.91\%$ OOD micro accuracy, confirming that both methods achieve comparable OOD performance under label-free fallback while \ours surpasses all three public RLVR checkpoints by $7.88\text{--}28.32\text{ pp}$ ($76.89\%$ vs.\ $48.57\%\text{--}69.01\%$) and all forward-only ZO baselines by $0.62\text{--}8.62\text{ pp}$ ($68.27\%\text{--}76.27\%$).

\section{Cross-Grid Scale Sensitivity and Contrastive Response Surface}
\label{sec:supp_scale}

\begin{table}[htbp]
  \centering
  \caption{\textbf{Spectral and subspace overlap sensitivity across $10$ scale factors $s \in [0.40, 0.96]$} on the $32$ unlabeled calibration diagrams ($r=32$). \textbf{Ov.\ Contra.$\%$} and \textbf{Ov.\ Pool.$\%$} measure Grassmannian overlap $\frac{1}{r}\|U(s)^\top U(0.75)\|_F^2$ with the $s=0.75$ basis for our contrastive operator ($M_\beta, \beta=0.35$) and pooled unpaired covariance ($C_{\mathrm{pool}}$); random rank-$32$ overlap is $1.25\%$ on 4B and $0.78\%$ on 8B.}
  \label{tab:supp_scale_sensitivity}
  \footnotesize
  \setlength{\tabcolsep}{4.8pt}
  \renewcommand{\arraystretch}{1.10}
  \begin{adjustbox}{max width=\textwidth}
  \begin{tabular}{cc ccccc cccc}
    \toprule
    \multirow{2}{*}{\textbf{Scale $s$}} & \multirow{2}{*}{\textbf{Moved $\%$}} & \multicolumn{5}{c}{\textbf{Qwen3-VL-4B} ($d=2560$, Rand.\ $=1.25\%$)} & \multicolumn{4}{c}{\textbf{Qwen3-VL-8B} ($d=4096$, Rand.\ $=0.78\%$)} \\
    \cmidrule(lr){3-7} \cmidrule(lr){8-11}
    & & $\Tr(C_{\mathrm{n}})/\Tr(C_{\mathrm{s}})$ & Self SNR & Ref$_{0.75}$ SNR & Ov.\ Contra.$\%$ & Ov.\ Pool.$\%$ & $\Tr(C_{\mathrm{n}})/\Tr(C_{\mathrm{s}})$ & Ref$_{0.75}$ SNR & Ov.\ Contra.$\%$ & Ov.\ Pool.$\%$ \\
    \midrule
    $0.40$ & 99.15 & 1.55 & 1.875 & 2.038 & 84.33 & 88.43 & 1.90 & 1.772 & 83.04 & 92.06 \\
    $0.50$ & 98.40 & 1.29 & 1.782 & 1.961 & 89.04 & 93.03 & 1.67 & 1.676 & 87.94 & 96.71 \\
    $0.60$ & 97.50 & 1.12 & 1.783 & 1.906 & 94.18 & 95.57 & 1.45 & 1.629 & 91.19 & 97.49 \\
    \rowcolor{oursrow}
    $0.65$ & 97.28 & 1.02 & 1.777 & 1.852 & \textbf{96.92} & \textbf{96.33} & 1.37 & 1.615 & \textbf{92.07} & \textbf{97.81} \\
    \rowcolor{oursrow}
    $0.70$ & 95.17 & 0.96 & 1.817 & 1.850 & \textbf{98.31} & \textbf{98.56} & 1.35 & 1.598 & \textbf{93.70} & \textbf{97.01} \\
    \rowcolor{oursrow}
    \textbf{0.75 (Ours)} & \textbf{93.81} & \textbf{0.90} & \textbf{1.840} & \textbf{1.840} & \textbf{100.0} & \textbf{100.0} & \textbf{1.30} & \textbf{1.580} & \textbf{100.0} & \textbf{100.0} \\
    \rowcolor{oursrow}
    $0.80$ & 90.26 & 0.84 & \textbf{1.858} & 1.788 & \textbf{97.87} & \textbf{99.30} & 1.23 & 1.535 & \textbf{94.52} & \textbf{98.61} \\
    \rowcolor{oursrow}
    $0.85$ & 82.78 & 0.77 & 1.793 & 1.669 & \textbf{95.97} & \textbf{98.50} & 1.17 & 1.487 & \textbf{93.45} & \textbf{98.40} \\
    $0.90$ & 69.25 & 0.66 & 1.669 & 1.506 & 93.30 & 98.62 & 1.05 & 1.397 & 91.00 & 98.22 \\
    $0.96$ & 33.42 & 0.43 & 1.627 & 1.376 & 88.58 & 97.85 & 0.78 & 1.305 & 86.86 & 97.43 \\
    \bottomrule
  \end{tabular}
  \end{adjustbox}
\end{table}

\cref{tab:supp_scale_sensitivity} evaluates how the rescaling factor $s$ affects token displacement, residual-to-signal trace ratio $\Tr(C_{\mathrm{n}})/\Tr(C_{\mathrm{s}})$, subspace SNR, and Grassmannian overlap with the default $s=0.75$ basis across $s \in [0.40, 0.96]$ on both Qwen3-VL-4B and Qwen3-VL-8B.
Across $s \in [0.65, 0.85]$, subspace overlap with $U(0.75)$ stays between $95.97\%$ and $100.0\%$ on 4B and between $92.07\%$ and $100.0\%$ on 8B (compared with $1.25\%$ and $0.78\%$ for random rank-$32$ subspaces).
For $s \ge 0.90$, cross-cell token movement drops ($93.81\% \to 33.42\%$ at $s=0.96$, leaving $66.6\%$ of tokens in their original cells), under-sampling boundary shift modes and reducing reference SNR ($1.840 \to 1.376$ on 4B; $1.580 \to 1.305$ on 8B).
For $s \le 0.50$, heavy downsampling blurs small symbols and raises $\Tr(C_{\mathrm{n}})/\Tr(C_{\mathrm{s}})$ to $1.55$ on 4B and $1.90$ on 8B.
Our default $s = 0.75$ lies within the stable middle range $[0.65, 0.85]$.

\begin{figure}[htbp]
  \centering
  \includegraphics[width=\textwidth]{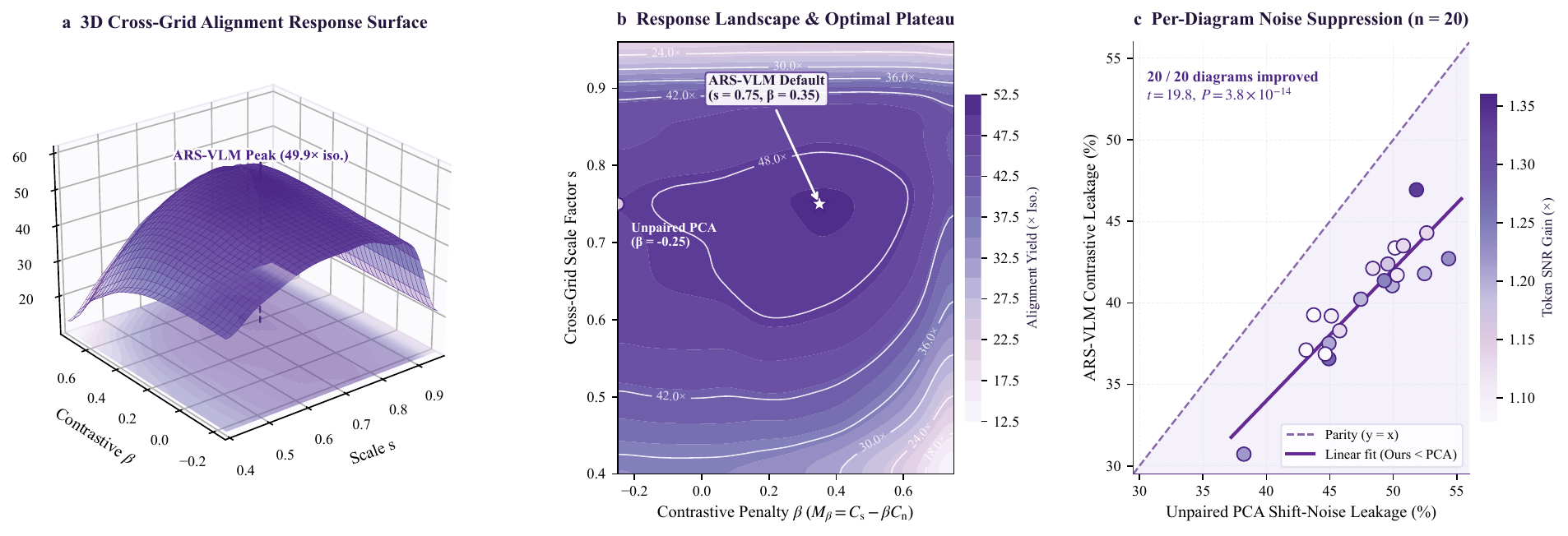}
  \caption{\textbf{Cross-grid spectral response surface and per-diagram shift-noise suppression on Qwen3-VL-8B ($r=32$).} \textbf{(a, b)} Held-out cross-grid alignment yield (\cref{eq:supp_align_yield}, relative to an isotropic rank-$32$ subspace) evaluated across $10$ rescaling factors $s \in [0.40, 0.96]$ and $13$ contrastive penalties $\beta \in [-0.25, 0.75]$ ($130$ grid configurations calibrated on $32$ unlabeled images and tested on $40$ held-out Geometry3K and MMK12-Dev diagrams); our default $(s=0.75, \beta=0.35)$ sits on the global peak ($49.9\times$ isotropic). \textbf{(c)} Per-diagram shift-noise leakage across $20$ held-out diagrams evaluated under the spatial token probe of \cref{fig:qualitative}, showing improvements in all $20/20$ diagrams over Unpaired PCA (paired $t = 19.8$, $P = 3.8 \times 10^{-14}$).}
  \label{fig:supp_3d_response}
\end{figure}

To jointly evaluate shared-signal retention, shift-noise suppression, boundary excitation, and symbol legibility across $(s, \beta)$, we measure the held-out cross-grid alignment yield on $40$ held-out diagrams ($24$ from Geometry3K and $16$ from MMK12-Dev):
\begin{equation}
\label{eq:supp_align_yield}
    \mathcal{Y}(s, \beta) = \frac{\big(\widetilde{\mathrm{Sig}}(s,\beta)\big)^2}{\widetilde{\mathrm{Noise}}(s,\beta)} \, \mu(s) \, \kappa(s)^2,
\end{equation}
where $\widetilde{\mathrm{Sig}}(s,\beta) = \Tr(U_{s,\beta}^\top C_{\mathrm{s}}^{\mathrm{eval}} U_{s,\beta}) / \Tr(C_{\mathrm{s}}^{\mathrm{eval}})$ and $\widetilde{\mathrm{Noise}}(s,\beta) = \Tr(U_{s,\beta}^\top C_{\mathrm{n}}^{\mathrm{eval}} U_{s,\beta}) / \Tr(C_{\mathrm{n}}^{\mathrm{eval}})$ denote the fractions of held-out cross-view signal and shift residual captured by $U_{s,\beta}$, $\mu(s)$ is the cross-cell token movement fraction (\textbf{Moved $\%$}), and $\kappa(s)$ is cross-view token cosine fidelity.
Normalized by the isotropic rank-$32$ baseline ($r/d = 32/4096$), \cref{fig:supp_3d_response}a,b shows a smooth dome peaking at $49.9\times$ right at our default $(s=0.75, \beta=0.35)$, whereas Unpaired PCA ($\beta = -0.25$, $M_{-0.25} = C_{\mathrm{pool}}$) lacks shift-noise suppression, $s \ge 0.90$ under-excites $28\text{px}$ cell boundaries, $s \le 0.50$ blurs fine glyphs, and $\beta \ge 0.60$ discards shared foreground variance.
Applying the spatial token probe of \cref{fig:qualitative} to $20$ held-out diagrams (\cref{fig:supp_3d_response}c), $U(0.75, 0.35)$ reduces shift-noise leakage ($30.7\%\text{--}46.9\%$ vs.\ $38.2\%\text{--}54.4\%$, paired $t = 19.8$, $P = 3.8 \times 10^{-14}$) and raises foreground cross-view cosine similarity ($0.668\text{--}0.870$ vs.\ $0.637\text{--}0.831$) in all $20/20$ cases.

\section{Per-Subject MMK12-Test Breakdown and Layer-Wise Dissection}
\label{sec:supp_mmk12}

\begin{table}[t]
  \centering
  \caption{\textbf{Per-subject accuracy, \texttt{\textbackslash boxed\{\}} format compliance, and mean generation length on held-out MMK12-Test.} \textbf{Top (8B \& Public RLVR, $N=2{,}000$, $500$/subject):} single-view $4{,}096$-token evaluation matching \cref{tab:main_8b}. \textbf{Bottom (4B MV-3, $N=400$ stratified, $100$/subject):} to resolve $2{,}048$-token single-view truncation in \cref{tab:ablations_4b} and test spatial invariance across $28\text{px}$ token grids, all 4B checkpoints from \cref{tab:ablations_4b} are evaluated at $4{,}096$ tokens via label-free 3-view cross-grid majority voting (\textbf{MV-3}: original $1.0\times$, aligned $0.75\times$, and half-cell shifted $+14\text{px}$ views; ties broken by shortest \texttt{\textbackslash boxed\{\}} trace). Following \cref{tab:main_8b}, first-order RL (\textbf{FO}) is listed as a reference and best forward-only \textbf{ZO} accuracy within each scale is in \textbf{bold}. The 8B Unpaired PCA run from \cref{tab:main_8b} is omitted because per-subject logs were not retained.}
  \label{tab:supp_mmk12_subjects}
  \footnotesize
  \setlength{\tabcolsep}{4.8pt}
  \renewcommand{\arraystretch}{1.10}
  \begin{adjustbox}{max width=\textwidth}
  \begin{tabular}{ll c cccc ccc}
    \toprule
    \textbf{Scale} & \textbf{Method} & \textbf{Mode} & \textbf{Math} & \textbf{Physics} & \textbf{Chemistry} & \textbf{Biology} & \textbf{Overall Acc.} & \textbf{Format $\%$} & \textbf{Mean Tokens} \\
    \midrule
    \multirow{15}{*}{\makecell[l]{8B\\(SV)}}
    & Pretrained Base (Step 0) & -- & 30.00 & 55.00 & 53.40 & 58.40 & 49.20 & 70.0 & 2050.6 \\
    \cmidrule{2-10}
    & MM-Eureka-Qwen-7B~\cite{meng2025mmeureka} & FO & 47.40 & 53.40 & 65.20 & 64.60 & 57.65 & 99.8 & 505.6 \\
    & NoisyRollout-MMK12-7B & FO & 38.00 & 52.40 & 61.80 & 64.00 & 54.05 & 99.0 & 509.5 \\
    & PRISM-DAPO-8B~\cite{wang2026prism} & FO & 16.40 & 4.60 & 4.40 & 3.40 & 7.20 & 16.2 & 2319.4 \\
    & Full-Param GRPO (Best)~\cite{shao2024deepseekmath} & FO & 29.80 & 61.20 & 62.60 & 67.00 & 55.15 & 84.5 & 1527.6 \\
    & Full-Param GRPO (Step 500)~\cite{shao2024deepseekmath} & FO & 26.60 & 56.40 & 62.60 & 67.40 & 53.25 & 83.5 & 1549.4 \\
    \cmidrule{2-10}
    & MeZO-RGE ($N=16$, Step 500)~\cite{malladi2023mezo} & ZO & 29.80 & 54.20 & 53.60 & 61.40 & 49.75 & 72.1 & 1945.6 \\
    & Vanilla ES (Best)~\cite{qiu2025esscale} & ZO & 22.00 & 54.00 & 56.00 & 61.00 & 48.25 & 79.7 & 1685.9 \\
    & Vanilla ES (Step 500)~\cite{qiu2025esscale} & ZO & 24.00 & 49.60 & 56.60 & 61.40 & 47.90 & 81.4 & 1583.3 \\
    & Vanilla ARS (Best, Step 90)~\cite{mania2018ars} & ZO & \textbf{36.60} & 56.80 & 56.20 & 63.60 & 53.30 & 77.5 & 1799.2 \\
    & Vanilla ARS (Step 500)~\cite{mania2018ars} & ZO & 33.20 & 55.40 & 53.60 & 57.40 & 49.90 & 82.9 & 1467.2 \\
    & Hard Subspace ($\rho=1.0$, Step 500) & ZO & 26.80 & 48.60 & 51.60 & 54.40 & 45.35 & 86.8 & 1365.7 \\
    & Random Orthonormal Subspace (Best) & ZO & 26.40 & 51.40 & 54.40 & 64.20 & 49.10 & 80.0 & 1653.7 \\
    & \cellcolor{oursrow}\ours \textbf{(Ours, Best)} & \cellcolor{oursrow}\textbf{ZO} & \cellcolor{oursrow}30.40 & \cellcolor{oursrow}57.20 & \cellcolor{oursrow}58.60 & \cellcolor{oursrow}\textbf{68.60} & \cellcolor{oursrow}53.70 & \cellcolor{oursrow}84.8 & \cellcolor{oursrow}1630.0 \\
    & \cellcolor{oursrow}\ours \textbf{(Ours, Step 500)} & \cellcolor{oursrow}\textbf{ZO} & \cellcolor{oursrow}31.60 & \cellcolor{oursrow}\textbf{57.60} & \cellcolor{oursrow}\textbf{59.60} & \cellcolor{oursrow}68.00 & \cellcolor{oursrow}\textbf{54.20} & \cellcolor{oursrow}85.5 & \cellcolor{oursrow}1562.0 \\
    \midrule
    \multirow{10}{*}{\makecell[l]{4B\\(MV-3)}}
    & Pretrained Base (4B, Step 0) & -- & 31.00 & 61.00 & 61.00 & 62.00 & 53.75 & 84.2 & 1496.6 \\
    \cmidrule{2-10}
    & Full-Param GRPO~\cite{shao2024deepseekmath} & FO & 28.00 & 53.00 & 68.00 & 76.00 & 56.25 & 90.8 & 1210.9 \\
    \cmidrule{2-10}
    & MeZO-RGE ($N=16$, Antithetic SPSA)~\cite{malladi2023mezo} & ZO & 30.00 & 59.00 & 57.00 & 69.00 & 53.75 & 83.2 & 1515.8 \\
    & Vanilla ARS~\cite{mania2018ars} & ZO & 29.00 & 50.00 & 64.00 & 67.00 & 52.50 & 91.8 & 1123.5 \\
    & Vanilla ES~\cite{qiu2025esscale} & ZO & \textbf{33.00} & 51.00 & \textbf{65.00} & \textbf{74.00} & 55.75 & 88.0 & 1226.8 \\
    & \cellcolor{oursrow}\ours \textbf{(Full Method)} & \cellcolor{oursrow}\textbf{ZO} & \cellcolor{oursrow}31.00 & \cellcolor{oursrow}\textbf{63.00} & \cellcolor{oursrow}\textbf{65.00} & \cellcolor{oursrow}67.00 & \cellcolor{oursrow}\textbf{56.50} & \cellcolor{oursrow}87.2 & \cellcolor{oursrow}1240.5 \\
    & \quad w/ Unpaired PCA & ZO & 29.00 & 59.00 & 64.00 & 69.00 & 55.25 \drop{-1.25} & 86.5 & 1270.2 \\
    & \quad w/ Random Orthonormal Subspace ($U_{\mathrm{rand}}$) & ZO & 24.00 & 56.00 & 64.00 & \textbf{74.00} & 54.50 \drop{-2.00} & 87.5 & 1265.4 \\
    & \quad w/o Dual-View Scoring (Single-View, $\lambda=0$) & ZO & 26.00 & 48.00 & 63.00 & 72.00 & 52.25 \drop{-4.25} & 86.2 & 1365.8 \\
    & \quad w/o Soft Prior (Hard Subspace $\rho=1.0$) & ZO & 29.00 & 61.00 & 62.00 & 66.00 & 54.50 \drop{-2.00} & 87.8 & 1299.1 \\
    \bottomrule
  \end{tabular}
  \end{adjustbox}
\end{table}

\paragraph{Per-Subject Breakdown (\cref{tab:supp_mmk12_subjects}).}
\cref{tab:supp_mmk12_subjects} reports per-subject accuracy across Mathematics, Physics, Chemistry, and Biology alongside \texttt{\textbackslash boxed\{\}} format compliance and mean completion length on MMK12-Test.
On Qwen3-VL-8B at step $500$, \ours raises format compliance from $70.0\%$ (Base) to $85.5\%$ (vs.\ $83.5\%$ for GRPO) and shortens average completion length by $23.8\%$ relative to Base ($1{,}562.0$ vs.\ $2{,}050.6$ tokens), achieving the highest Biology accuracy across all first-order, public RLVR, and forward-only checkpoints ($68.00\%$ at step $500$ and $68.60\%$ at Best) and leading all ZO baselines in Physics ($57.60\%$), Chemistry ($59.60\%$), and overall accuracy ($54.20\%$).
On Qwen3-VL-4B, single-view decoding at $2{,}048$ tokens (\cref{tab:ablations_4b}) leaves over one-third of MMK12-Test chains truncated before closing \texttt{\textbackslash boxed\{\}}; evaluating all \cref{tab:ablations_4b} checkpoints at $4{,}096$ tokens under label-free 3-view cross-grid majority voting (\textbf{MV-3}: $1.0\times$, $0.75\times$, and $+14\text{px}$ half-cell shifted views) raises \ours format compliance to $87.2\%$ and reaches $56.50\%$ overall accuracy ($63.00\%$ in Physics), outperforming first-order Full-Param GRPO ($56.25\%$), all ZO baselines ($52.50\%\text{--}55.75\%$), and all four single-component ablations ($-1.25$ to $-4.25\text{ pp}$, with removing dual-view scoring causing the largest $-4.25\text{ pp}$ drop).

\begin{table}[htbp]
  \centering
  \caption{\textbf{Cumulative weight Frobenius drift $\|\theta_t - \theta_0\|_F$ on Qwen3-VL-8B} across steps $100\to 500$ ($1{,}000$ for ES), with shallow (\texttt{0..17}) and deep (\texttt{18..35}) decoder-layer norms at the final step.}
  \label{tab:supp_drift}
  \small
  \setlength{\tabcolsep}{8.0pt}
  \renewcommand{\arraystretch}{1.12}
  \begin{adjustbox}{max width=\textwidth}
  \begin{tabular}{l ccccc cc}
    \toprule
    \textbf{Method} & \textbf{100} & \textbf{200} & \textbf{300} & \textbf{400} & \textbf{500} & \textbf{L0--17} & \textbf{L18--35} \\
    \midrule
    Vanilla ARS & 132.0 & 192.1 & 233.5 & 267.5 & 303.9 & 191.3 & 191.9 \\
    Hard ($\rho=1.0$) & 150.5 & 213.4 & 263.2 & 297.1 & 303.0 & 178.9 & 191.7 \\
    Vanilla ES & 91.0 & 128.7 & 157.5 & 181.8 & 203.1 & 128.0 & 128.1 \\
    Vanilla ES (1k) & -- & -- & -- & -- & 286.4 & 180.5 & 180.7 \\
    \rowcolor{oursrow}
    \ours & 140.5 & 187.5 & 210.5 & 219.1 & \textbf{223.3} & \textbf{131.7} & 145.7 \\
    \bottomrule
  \end{tabular}
  \end{adjustbox}
\end{table}

\begin{figure}[htbp]
  \centering
  \includegraphics[width=\textwidth]{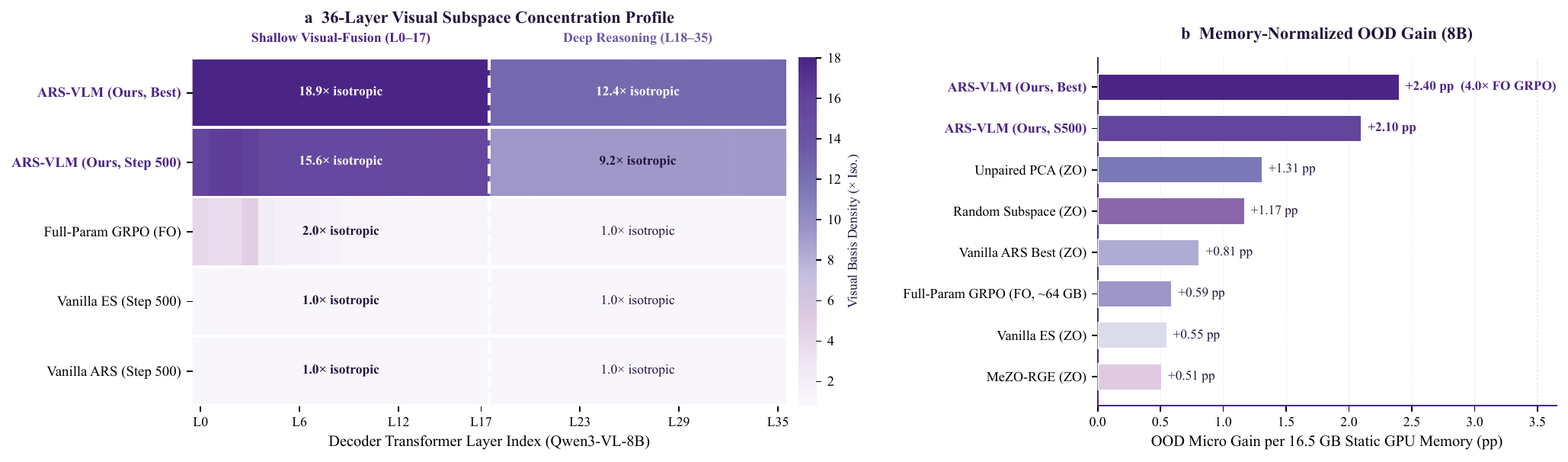}
  \caption{\textbf{Layer-wise visual subspace concentration and memory-normalized OOD gain on Qwen3-VL-8B.} \textbf{(a)} Cumulative weight-update energy captured by the rank-$32$ contrastive visual basis $U$ across all $36$ decoder layers (expressed as multiples of the isotropic baseline $r/d = 0.78\%$). \ours concentrates updates in shallow visual-fusion layers (\texttt{L0--17}, $18.9\times$ at Best and $15.6\times$ at step $500$) while relaxing deeper reasoning layers (\texttt{L18--35}, $12.4\times$ and $9.2\times$), whereas unconstrained ZO and first-order GRPO remain near-isotropic ($1.0\times\text{--}2.0\times$). \textbf{(b)} Single-view 7-OOD micro accuracy gain over Base normalized per $16.5\text{ GB}$ of static GPU memory (\cref{tab:main_8b,tab:supp_memory}), where \ours achieves $4.0\times$ the memory-normalized OOD gain of Full-Param GRPO ($+2.40$ vs.\ $+0.59\text{ pp}$ per $16.5\text{ GB}$).}
  \label{fig:supp_layer_profile}
\end{figure}

\paragraph{Layer-Wise Weight Drift and Subspace Profile (\cref{tab:supp_drift,fig:supp_layer_profile}).}
\cref{tab:supp_drift} reports cumulative Frobenius norms $\|\theta_t - \theta_0\|_F$ from step $100$ to step $500$ on Qwen3-VL-8B, split at step $500$ into shallow visual-fusion layers (\texttt{layers.0..17}) and deeper reasoning layers (\texttt{layers.18..35}).
Without step-size annealing or a reward-dispersion floor, Vanilla ARS and Hard Subspace ($\rho=1.0$) exceed $303$ in total Frobenius norm, coinciding with the OOD drops in \cref{tab:main_8b}.
\ours slows parameter movement between steps $400$ and $500$ ($219.1 \to 223.3$) and reduces shallow-layer drift by $31.1\%$ relative to Vanilla ARS ($131.7$ vs.\ $191.3$), matching the stabilization of Schulman policy KL in \cref{fig:dynamics}c.

\cref{fig:supp_layer_profile}a dissects the resulting weight updates $\Delta\theta = \theta_t - \theta_0$ across all $36$ decoder blocks by measuring the fraction of update Frobenius energy aligned with $U$ relative to random rank-$32$ density ($r/d = 0.78\%$).
Because Stage~2 retains the $\rho_{\mathrm{soft}} = 0.25$ prior exclusively on shallow layers ($\ell < 18$) after the $20$-step global warmup, \ours exhibits a clear step transition at the \texttt{L17}/\texttt{L18} boundary ($18.9\times \to 12.4\times$ at Best; $15.6\times \to 9.2\times$ at step $500$), preserving visual-token grounding in early blocks while leaving deep blocks free for multi-step STEM deduction.
Combining the single-view OOD gains of \cref{tab:main_8b} with the static memory footprint of \cref{tab:supp_memory}, \cref{fig:supp_layer_profile}b shows that \ours delivers $+2.40\text{ pp}$ (Best) and $+2.10\text{ pp}$ (step $500$) per $16.5\text{ GB}$ worker GPU, outperforming all forward-only ZO baselines as well as Full-Param GRPO ($+0.59\text{ pp}$ per $16.5\text{ GB}$ at ${\sim}64\text{ GB/GPU}$, a $4.0\times$ efficiency ratio).

\section{Cross-Benchmark OOD Landscape and Component Necessity}
\label{sec:supp_ood_landscape}

\begin{figure}[htbp]
  \centering
  \includegraphics[width=\textwidth]{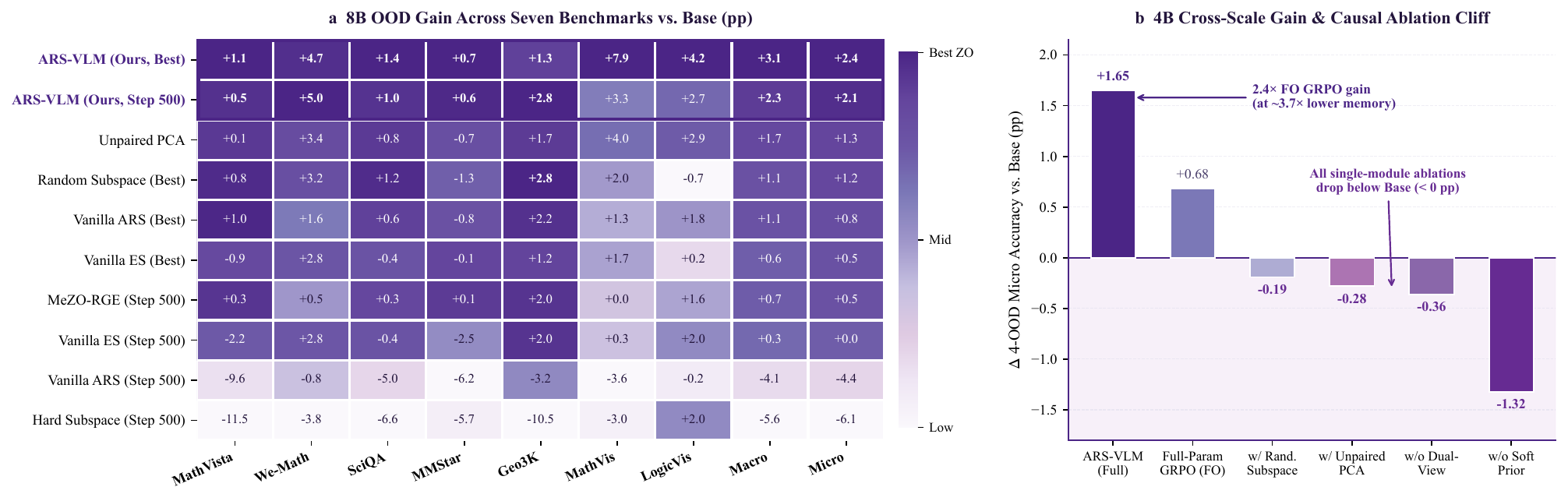}
  \caption{\textbf{Cross-benchmark OOD generalization landscape (8B) and causal component ablation cliff (4B).} \textbf{(a)} Per-benchmark OOD accuracy differences ($\text{pp}$ vs.\ Pretrained Base) across all seven OOD suites plus Macro and Micro summaries on Qwen3-VL-8B (\cref{tab:main_8b}; cell shading normalized within each column across ZO checkpoints). \ours (boxed top rows) sustains positive transfer across all seven benchmarks at both Best ($+2.4\text{ pp}$ Micro) and step $500$ ($+2.1\text{ pp}$ Micro). \textbf{(b)} Qwen3-VL-4B 4-OOD micro accuracy differences relative to Base ($71.77\%$, \cref{tab:ablations_4b}): full \ours achieves $+1.65\text{ pp}$ ($2.4\times$ Full-Param GRPO at ${\sim}3.7\times$ lower static memory on 4B), whereas removing any single component drops OOD accuracy below Base.}
  \label{fig:supp_ood_heatmap}
\end{figure}

\cref{fig:supp_ood_heatmap} synthesizes how each algorithmic component contributes to out-of-distribution transfer across model scales.
On Qwen3-VL-8B (\cref{fig:supp_ood_heatmap}a), isotropic Vanilla ARS peaks early before falling $4.4\text{ pp}$ below Base by step $500$ ($-5.21\text{ pp}$ from its step-$90$ peak, reaching $-9.6\text{ pp}$ on MathVista and $-6.2\text{ pp}$ on MMStar), and hard rank-$32$ projection ($\rho = 1.0$) falls $6.1\text{ pp}$ below Base Micro ($-11.5\text{ pp}$ on MathVista and $-10.5\text{ pp}$ on Geometry3K).
In contrast, \ours maintains positive gains across every single OOD benchmark at both its validation-selected checkpoint (up to $+7.9\text{ pp}$ on MathVision, $+4.7\text{ pp}$ on We-Math, and $+4.2\text{ pp}$ on LogicVista) and step $500$.
On Qwen3-VL-4B (\cref{fig:supp_ood_heatmap}b), full \ours improves 4-OOD micro accuracy by $+1.65\text{ pp}$ over Base ($2.4\times$ the $+0.68\text{ pp}$ gain of first-order GRPO), whereas replacing the paired contrastive basis with $U_{\mathrm{rand}}$ ($-0.19\text{ pp}$) or Unpaired PCA ($-0.28\text{ pp}$), removing dual-view scoring ($-0.36\text{ pp}$), or removing the Stage~2 soft release ($-1.32\text{ pp}$) uniformly degrades OOD performance below the pretrained base model.

\section{Experimental Setup and Hyperparameter Summary}
\label{sec:supp_repro}

\begin{table}[htbp]
  \centering
  \caption{\textbf{Summary of experimental setup, optimization hyperparameters, and evaluation protocols} across Qwen3-VL-8B and Qwen3-VL-4B.}
  \label{tab:supp_hyperparams}
  \small
  \setlength{\tabcolsep}{9.0pt}
  \renewcommand{\arraystretch}{1.14}
  \begin{adjustbox}{max width=\textwidth}
  \begin{tabular}{ll}
    \toprule
    \textbf{Component / Method} & \textbf{Configuration} \\
    \midrule
    \multicolumn{2}{l}{\emph{\ours Hyperparameters (Identical Across 4B and 8B)}} \\
    Calibration set $M_{\mathrm{cal}}$ & $32$ unlabeled training diagrams (${\sim}30\text{s}$) \\
    Cross-grid aligned view & Scale $s=0.75$, $28\text{px}$-aligned canvas, $1.08\times$ jitter \\
    Contrastive basis $U$ & $M_\beta = C_{\mathrm{s}} - \beta C_{\mathrm{n}}$, $\beta = 0.35$, rank $r = 32$ \\
    Stage 1 warmup ($t \le 20$) & $\rho_{t,\ell} = 1.0$ across all $36$ decoder layers \\
    Stage 2 soft prior ($t > 20$) & $\rho_{\mathrm{soft}} = 0.25$ ($\ell < 18$); $\rho = 0$ ($\ell \ge 18$) \\
    ARS population \& scale & $N=16$ antithetic pairs, top-$k=8$, $\sigma_{\mathrm{noise}}=10^{-3}$ \\
    Step size \& normalization & $\alpha_0 = 10^{-4}$, cosine floor $0.25\alpha_0$, $\sigma_{\mathrm{eff}} = \max(\sigma_{\mathrm{all}}, \sigma_{\mathrm{sel}})$ \\
    Dual-view penalty $\lambda$ & $\lambda = 0.1$ on $C_i$ (\cref{eq:inconsistency}); $4NB = 1{,}024$ rollouts/step \\
    \midrule
    \multicolumn{2}{l}{\emph{Hardware, Baselines, and Evaluation Settings}} \\
    Compute hardware & $8 \times \text{NVIDIA A100}$ ($80\text{ GB}$) GPUs for all experiments \\
    Full-Param GRPO (FO) & $G=8$, $\text{lr}=10^{-6}$, AdamW, $\beta_{\mathrm{KL}}=0.01$; $BG=128$ rollouts/step \\
    Vanilla ES (ZO) & $P=30$, $\sigma=10^{-3}$, $\alpha=5\times 10^{-4}$, $z$-scored rewards \\
    Vanilla ARS (ZO) & $N=16$, $k=8$, $\sigma_{\mathrm{noise}}=10^{-3}$, $\alpha=10^{-4}$; $2NB=512$ rollouts/step \\
    MeZO-RGE (ZO) & $N=16$ antithetic pairs, $\sigma_{\mathrm{noise}}=10^{-3}$, $\eta=10^{-6}$; $2NB=512$/step \\
    MMK12 batch \& horizon & $B=16$; $500$ steps on both 4B and 8B (including all ablations) \\
    Pure-text forgetting suite & $4$ tasks $\times$ $4$ methods, $150$ steps/run, eval every $10$ steps \\
    Evaluation decoding & Greedy (\texttt{temperature=0}); max $4{,}096$ tok (\cref{tab:main_8b}, \cref{tab:supp_mmk12_subjects} MV-3), $2{,}048$ tok (\cref{tab:ablations_4b}) \\
    \bottomrule
  \end{tabular}
  \end{adjustbox}
\end{table}

\cref{tab:supp_hyperparams} summarizes the hardware configuration, optimization hyperparameters, per-step rollout counts, and decoding limits used throughout the paper.
All training, ablation, and evaluation runs use $8 \times \text{NVIDIA A100}$ ($80\text{ GB}$) GPUs, and all \ours hyperparameters in \cref{tab:supp_hyperparams} are shared across model scales and task suites without task-specific retuning.

\section{Prompt Templates and Answer Verification}
\label{sec:supp_prompts}

\begin{table}[htbp]
  \centering
  \caption{\textbf{System prompts and user-turn templates used across multimodal and text-only experiments.} All inputs are formatted with the native Qwen3-VL chat template (\texttt{add\_generation\_prompt=True}); multimodal inputs attach the RGB diagram (original or $0.75\times$ aligned view) before the user text.}
  \label{tab:supp_prompts}
  \setlength{\tabcolsep}{5.0pt}
  \renewcommand{\arraystretch}{1.18}
  \footnotesize
  \begin{tabular}{>{\raggedright\arraybackslash}p{0.20\textwidth} >{\raggedright\arraybackslash}p{0.34\textwidth} >{\raggedright\arraybackslash}p{0.39\textwidth}}
    \toprule
    \textbf{Suite / Benchmark} & \textbf{System Prompt} & \textbf{User-Turn Template} \\
    \midrule
    \textbf{MMK12} (Train, Dev, Test) \& Open-Ended Vision OOD (\textbf{Geometry3K}, \textbf{LogicVista}, \textbf{MathVista}, \textbf{MathVision} free-form) &
    \texttt{You are solving a problem shown in the image. Read the image and the question. On the last line, put only the final answer in \textbackslash boxed\{\}.} &
    \texttt{[Image] \{question\}} \newline (For MathVista, \texttt{\{question\}} is the official \texttt{query} field.) \\
    \midrule
    Multiple-Choice Vision OOD (\textbf{MMStar}, \textbf{We-Math}, \textbf{ScienceQA-IMG}, \textbf{MathVision} multi-choice) &
    \texttt{You are solving a problem shown in the image. Read the image and the question. On the last line, put only the final answer in \textbackslash boxed\{\}.} &
    \textbf{MMStar:} \texttt{[Image] \{question\}\textbackslash nPlease put the correct option letter (A, B, C, or D) in \textbackslash boxed\{\}.} \newline
    \textbf{We-Math:} \texttt{[Image] \{question\}\textbackslash nOptions: \{options\}\textbackslash nPlease put the correct option letter (e.g., A, B, C, D, or E) in \textbackslash boxed\{\}.} \newline
    \textbf{ScienceQA-IMG:} \texttt{[Image] Context: \{hint\}\textbackslash nQuestion: \{question\}\textbackslash nOptions:\textbackslash n(A) ...\textbackslash nPlease put the correct option letter (e.g., A, B, C, or D) in \textbackslash boxed\{\}.} \\
    \midrule
    \textbf{Countdown} (Pure-Text RL) &
    \texttt{You are a helpful assistant. Solve the arithmetic puzzle step by step, and put your final equation inside \textbackslash boxed\{...\} or <answer>...</answer> tags.} &
    \texttt{Using the numbers \{nums\}, create an equation that equals \{target\}. You can use basic arithmetic operations (+, -, *, /) and parentheses, and each number must be used exactly once. Show your reasoning and put your final equation inside \textbackslash boxed\{...\}.} \\
    \midrule
    \textbf{GSM8K}, \textbf{MATH-500}, \textbf{OlympiadBench} (Pure-Text RL) &
    \texttt{Solve the following problem step by step. On the last line, put only the final answer in \textbackslash boxed\{\}.} &
    \texttt{\{problem\}} \\
    \midrule
    \textbf{HellaSwag} (Language Prior Retention) &
    \texttt{Read the context carefully and select the most plausible continuation or answer. Put only the final option letter (A, B, C, or D) in \textbackslash boxed\{\}.} &
    \texttt{Activity: \{activity\_label\}\textbackslash nContext: \{ctx\}\textbackslash n\textbackslash nWhich continuation makes the most sense?\textbackslash nA. \{ending\_0\} ... D. \{ending\_3\}\textbackslash n\textbackslash nPut only the option letter (A, B, C, or D) in \textbackslash boxed\{\}.} \\
    \bottomrule
  \end{tabular}
\end{table}

\paragraph{Prompt Formatting (\cref{tab:supp_prompts}).}
\cref{tab:supp_prompts} lists the exact system instructions and user-turn templates used across all training and evaluation suites.
All multimodal runs share a single system prompt requiring the final answer inside \texttt{\textbackslash boxed\{\}}, without few-shot exemplars or benchmark-specific chain-of-thought demonstrations.
For dual-view training and DV-Fallback, the identical user text and system prompt are paired with either the original RGB image $I_{\mathrm{o}}$ or the $0.75\times$ centered-canvas view $I_{\mathrm{a}}$.

\paragraph{Deterministic Reward and Answer Verification.}
During MMK12 post-training (both GRPO and all forward-only ZO methods), the verifier extracts the last balanced \texttt{\textbackslash boxed\{...\}} span from the model output; responses without a closed \texttt{\textbackslash boxed\{\}} receive reward $0$.
Extracted strings and reference answers undergo standard LaTeX normalization (stripping math delimiters, whitespace, \texttt{\textbackslash left}/\texttt{\textbackslash right}, \texttt{\textbackslash text}, normalizing \texttt{\textbackslash dfrac}/\texttt{\textbackslash tfrac} to \texttt{\textbackslash frac}, and removing trailing unit suffixes such as degrees or metric units).
A prediction receives reward $1$ if it matches the normalized reference string, agrees numerically within $\max(10^{-3}, 10^{-3}|a|)$ when both parse as finite floats, or matches the reference option letter on multiple-choice items.

In the pure-text forgetting suite (\cref{tab:reasoning_4b}), GSM8K, MATH-500, OlympiadBench, and HellaSwag use the same \texttt{\textbackslash boxed\{\}} equivalence grader. Following Abdi~\etal~\cite{abdi2026forgetting}, Countdown parses the arithmetic expression inside \texttt{\textbackslash boxed\{...\}} or \texttt{<answer>...</answer>}, verifies that the integer multiset matches the input numbers, evaluates the expression via an abstract syntax tree, and assigns $0.1$ for valid syntax plus $0.9$ for matching the target value during training (reporting exact $0/1$ target accuracy at evaluation).
During $4{,}096$-token benchmark evaluation (\cref{tab:main_8b}), every checkpoint is scored by the same \texttt{\textbackslash boxed\{\}} verifier first; if no \texttt{\textbackslash boxed\{\}} span is present, a uniform tail parser checks the final lines for explicit answer cues (e.g., ``\texttt{Final answer:}'', ``\texttt{The answer is}'', or a standalone final option letter or number) before grading.

\section{Qualitative and Frontier-Judge Evaluation of Visually Aligned Reasoning}
\label{sec:supp_qualitative_cases}

\begin{figure}[t]
  \centering
  \setlength{\tabcolsep}{3.5pt}
  \renewcommand{\arraystretch}{1.18}
  \scriptsize\linespread{1.18}\selectfont
  \begin{tabular}{>{\raggedright\arraybackslash}p{0.184\textwidth} >{\raggedright\arraybackslash}p{0.174\textwidth} >{\raggedright\arraybackslash}p{0.190\textwidth} >{\raggedright\arraybackslash}p{0.180\textwidth} >{\columncolor{oursrow}\raggedright\arraybackslash}p{0.190\textwidth}}
    \toprule
    \textbf{Problem \& Diagram} &
    \textbf{Pretrained Base (Step 0)} &
    \textbf{Full-Param GRPO (FO)} &
    \textbf{Vanilla ARS (ZO)} &
    \textbf{\ours (Ours, ZO)} \\
    \midrule
    \textbf{(a) Geometry3K} (Ref: $178.2$)\par\vspace{1.5pt}
    \centering\includegraphics[width=0.94\linewidth,height=1.40cm,keepaspectratio]{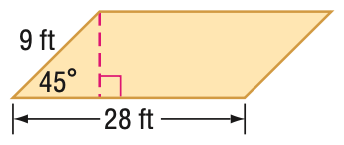}\par\vspace{1.5pt}
    \raggedright\emph{Q:} Find the area of the parallelogram. Round to the nearest tenth if necessary. &
    ``We are given a parallelogram with: a base of $28\text{ ft}$, an angle of $45^\circ$, \textcolor{dropred}{a height of $9\text{ ft}$ (shown as a perpendicular dashed line from the top side to the base)}. [\dots] $\text{Area} = 28 \times 9 = 252\text{ ft}^2$. [\dots] \textcolor{dropred}{The $45^\circ$ angle is likely there to distract} [\dots]''\par\vspace{1.5pt}
    \textcolor{dropred}{\textbf{Pred:}} $\boxed{252.0}$ {\color{dropred}($\times$, $245$ tok)} &
    ``We are given a parallelogram with a base of $28\text{ ft}$ and an angle of $45^\circ$. \textcolor{dropred}{The height is given as $9\text{ ft}$} [\dots] $\text{Area} = 28 \times 9 = 252\text{ ft}^2$. [\dots] \textcolor{dropred}{since the height is already given as $9\text{ ft}$, we don't need to calculate it from the angle.}''\par\vspace{1.5pt}
    \textcolor{dropred}{\textbf{Pred:}} $\boxed{252}$ {\color{dropred}($\times$, Best $275$ tok; {\color{gainteal}S500: $178.1$, $\checkmark$})} &
    ``Base $= 28\text{ ft}$, \textcolor{dropred}{Height $= 9\text{ ft}$} [\dots] \textcolor{dropred}{the height is already given as $9\text{ ft}$, and it is drawn perpendicular to the base} (indicated by the right angle symbol). [\dots] $\text{Area} = 28\text{ ft} \times 9\text{ ft} = 252\text{ ft}^2$ [\dots]''\par\vspace{1.5pt}
    \textcolor{dropred}{\textbf{Pred:}} $\boxed{252.0}$ {\color{dropred}($\times$; Best $215$ / S500 $155$ tok)} &
    ``From the image, we are given: the base of the parallelogram is $28\text{ ft}$; \textcolor{gainteal}{\textbf{the slanted side is $9\text{ ft}$, and the angle between the base and this side is $45^\circ$}}; the height is [\dots] perpendicular to the base. [\dots] $\sin(45^\circ) = \frac{h}{9}$ [\dots] $h \approx 6.3639\text{ ft}$ [\dots] $\text{Area} = 28 \times 6.3639 \approx 178.2\text{ ft}^2$.''\par\vspace{1.5pt}
    \textcolor{gainteal}{\textbf{Pred:}} $\boxed{178.2}$ {\color{gainteal}($\checkmark$; Best $481$ / S500 $368$ tok)} \\
    \midrule
    \textbf{(b) Geometry3K} (Ref: $7.2$)\par\vspace{1.5pt}
    \centering\includegraphics[width=0.84\linewidth,height=1.45cm,keepaspectratio]{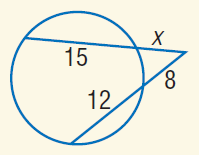}\par\vspace{1.5pt}
    \raggedright\emph{Q:} Find $x$ to the nearest tenth. Assume that segments that appear to be tangent are tangent. &
    ``One secant has \textcolor{dropred}{a segment of length $15$ outside the circle and a segment of length $12$ inside} [\dots] The other secant has $8$ outside [\dots] and \textcolor{dropred}{an unknown segment of length $x$ inside} [\dots] $15 \times (15 + 12) = 8 \times (8 + x)$ [\dots] $x = 42.625$.''\par\vspace{1.5pt}
    \textcolor{dropred}{\textbf{Pred:}} $\boxed{42.6}$ {\color{dropred}($\times$, $1{,}913$ tok)} &
    ``\textcolor{dropred}{One secant has a segment of length $15$ outside the circle and a chord of length $12$ inside the circle.} The other secant has $8$ outside [\dots] and \textcolor{dropred}{an unknown segment of length $x$ outside} [\dots] $12 \times 27 = x \times (x + 8)$ [\dots]''\par\vspace{1.5pt}
    \textcolor{dropred}{\textbf{Pred:}} $\boxed{14.4}$ {\color{dropred}($\times$, Best $1{,}696$ tok; S500: $\boxed{8.8}$, $\times$)} &
    ``The diagram shows: a secant segment [\dots] inside the circle ($12$) and [\dots] outside ($x$); \textcolor{dropred}{a tangent segment of length $8$ from the same external point}. [\dots] $8^2 = (x + 12) \times x$ [\dots]''\par\vspace{1.5pt}
    \textcolor{dropred}{\textbf{Pred:}} $\boxed{4.0}$ {\color{dropred}($\times$, Best $605$ tok; S500: $4{,}096$-tok loop)} &
    ``\textcolor{gainteal}{\textbf{One secant has a segment of length $15$ inside the circle and extends to a point outside with length $x$. The other secant has $12$ inside and extends to the same external point with length $8$.}} [\dots] $x(x + 15) = 8 \cdot 20$ [\dots] $x \approx 7.205$.''\par\vspace{1.5pt}
    \textcolor{gainteal}{\textbf{Pred:}} $\boxed{7.2}$ {\color{gainteal}($\checkmark$; Best $542$ / S500 $645$ tok)} \\
    \midrule
    \textbf{(c) We-Math} (Ref: \texttt{A}, $40^\circ$)\par\vspace{1.5pt}
    \centering\includegraphics[width=0.96\linewidth,height=1.40cm,keepaspectratio]{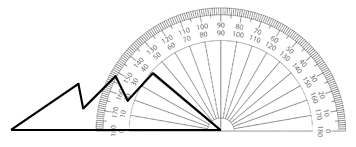}\par\vspace{1.5pt}
    \raggedright\emph{Q:} Using a protractor to measure one angle of the triangular glass fragment, the measured angle is $()^\circ$? A.~$40$; B.~$50$; C.~$60$; D.~None. &
    ``The base of the triangle aligns with the $0^\circ$ mark on the protractor. [\dots] Looking at the protractor scale, \textcolor{dropred}{the slanted side of the triangle points to the $50^\circ$ mark}.''\par\vspace{1.5pt}
    \textcolor{dropred}{\textbf{Pred:}} $\boxed{\text{B}}$ {\color{dropred}($\times$, $154$ tok)} &
    ``The baseline of the protractor is aligned with one side of the angle (the bottom side of the triangle). [\dots] \textcolor{dropred}{The other side of the angle (the slanted side) points to the $50^\circ$ mark on the protractor.}''\par\vspace{1.5pt}
    \textcolor{dropred}{\textbf{Pred:}} $\boxed{\text{B}}$ {\color{dropred}($\times$; Best $198$ / S500 $192$ tok)} &
    ``The ray on the left side points to $40^\circ$ on the outer scale [\dots] \textcolor{dropred}{The ray on the right side points to $90^\circ$ on the outer scale. So the angle is $90^\circ - 40^\circ = 50^\circ$.}''\par\vspace{1.5pt}
    \textcolor{dropred}{\textbf{Pred:}} $\boxed{\text{B}}$ {\color{dropred}($\times$; Best $1{,}172$ / S500 $187$ tok)} &
    ``\textcolor{gainteal}{\textbf{The base of the triangle (the bottom side) aligns with the $0^\circ$ mark}} [\dots] \textcolor{gainteal}{\textbf{we read the scale that increases from $0^\circ$. The ray forming the angle goes up to the $40^\circ$ mark}} on the protractor.''\par\vspace{1.5pt}
    \textcolor{gainteal}{\textbf{Pred:}} $\boxed{\text{A}}$ {\color{gainteal}($\checkmark$; Best $222$ / S500 $248$ tok)} \\
    \midrule
    \textbf{(d) We-Math} (Ref: \texttt{A}, $40$)\par\vspace{1.5pt}
    \centering\includegraphics[width=0.96\linewidth,height=1.40cm,keepaspectratio]{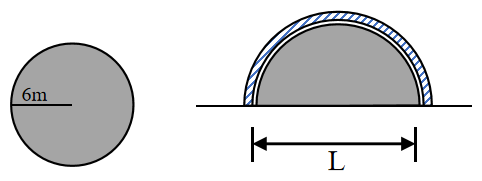}\par\vspace{1.5pt}
    \raggedright\emph{Q:} The diameter $L$ of the semicircle is close to the circumference of the small circle. Closest length to $L$? ($\pi=3.14$) A.~$40$; B.~$30$; C.~$50$; D.~None. &
    ``We are given \textcolor{dropred}{a small circle with diameter $6\text{ m}$}. [\dots] $C = \pi \times \text{diameter} = 3.14 \times 6 = 18.84\text{ m}$ [\dots] $18.84$ is closest to $30$.''\par\vspace{1.5pt}
    \textcolor{dropred}{\textbf{Pred:}} $\boxed{\text{B}}$ {\color{dropred}($\times$, $241$ tok)} &
    ``We are given \textcolor{dropred}{a small circle with diameter $6\text{ m}$}. [\dots] $C = 3.14 \times 6 = 18.84\text{ m}$ [\dots] $18.84$ is closest to $30$ among the given options.''\par\vspace{1.5pt}
    \textcolor{dropred}{\textbf{Pred:}} $\boxed{\text{B}}$ {\color{dropred}($\times$, Best $254$ tok; S500: $\boxed{\text{D}}$, $\times$)} &
    ``We are given \textcolor{dropred}{a small circle with diameter $6\text{ m}$}. [\dots] $C = 3.14 \times 6 = 18.84\text{ m}$ [\dots] \textcolor{dropred}{none of the options A, B, C are close to $18.84$.}''\par\vspace{1.5pt}
    \textcolor{dropred}{\textbf{Pred:}} $\boxed{\text{D}}$ {\color{dropred}($\times$; Best $445$ / S500 $231$ tok)} &
    ``We are given \textcolor{gainteal}{\textbf{a small circle with radius $6\text{ m}$}}. [\dots] $C = 2\pi r$ [\dots] \textcolor{gainteal}{\textbf{$C = 2 \times 3.14 \times 6 = 37.68\text{ m}$}} [\dots] $L \approx 37.68\text{ m}$ [\dots] $37.68$ is closest to $40$.''\par\vspace{1.5pt}
    \textcolor{gainteal}{\textbf{Pred:}} $\boxed{\text{A}}$ {\color{gainteal}($\checkmark$; Best $251$ / S500 $251$ tok)} \\
    \bottomrule
  \end{tabular}
  \caption{\textbf{Qualitative comparison of multimodal reasoning rollouts across Base, first-order Full-Param GRPO, isotropic Vanilla ARS, and \ours on Qwen3-VL-8B-Instruct} under the $4{,}096$-token protocol of \cref{tab:main_8b} and prompt templates of \cref{tab:supp_prompts}. Excerpts quote verbatim generations from the best validation checkpoint with routine intermediate algebra condensed as $[\dots]$ (S500 denotes the step-$500$ checkpoint, which yields the same boxed prediction unless noted); \textcolor{dropred}{red text} marks visual mis-grounding of diagram primitives, and \textcolor{gainteal}{\textbf{green bold text}} marks accurate visual alignment.}
  \label{fig:supp_qualitative_cases}
\end{figure}

While \cref{fig:qualitative} in the main text examines visual alignment at the token-representation level, \cref{fig:supp_qualitative_cases} compares end-to-end reasoning traces produced by Pretrained Base, first-order Full-Param GRPO, isotropic Vanilla ARS, and \ours on Qwen3-VL-8B-Instruct.
Across all four examples, the underlying mathematical formulas (parallelogram area, secant power of a point, protractor angle reading, and circle circumference) are already known to all checkpoints; errors arise instead when visual primitives are mis-bound to symbolic variables before algebraic deduction begins, even after full-parameter first-order RL on single-view trajectories.

Specifically, in \cref{fig:supp_qualitative_cases}a, Base, validation-selected Full-Param GRPO, and Vanilla ARS all attach the $9\text{ ft}$ label to the adjacent dashed perpendicular altitude rather than the slanted edge, treating the $45^\circ$ angle as unused context.
In \cref{fig:supp_qualitative_cases}b, Base swaps interior chords and exterior secant segments, Full-Param GRPO splices segments from two different secants ($15$ and $12$) at both Best and step $500$, and Vanilla ARS misinterprets the exterior segment of length $8$ as a tangent before degenerating into a $4{,}096$-token loop at step $500$.
In \cref{fig:supp_qualitative_cases}c,d, Base, Full-Param GRPO, and Vanilla ARS alike misread fine radial tick alignment on the dual-scale protractor ($50^\circ$ vs.\ $40^\circ$) and mistake a center-terminating radius segment ($r=6\text{ m}$) for a full diameter at both checkpoints.
By steering shallow-layer perturbations along cross-grid consistent visual directions and penalizing dual-view rollout disagreement, \ours accurately grounds edge labels, boundary intersections, scale ticks, and segment endpoints at both its best validation checkpoint and step $500$.

\begin{table}[t]
  \centering
  \caption{\textbf{Blind visual-reasoning alignment audit across $205$ stratified OOD diagrams} scored by \texttt{Claude Fable 5.2} (\texttt{high} thinking effort) with anonymized, permuted candidates (\texttt{GPT-6.2 Astra} cross-judge validation in text; prompt in \cref{tab:supp_judge_prompt}). Best \textbf{ZO} in \textbf{bold}.}
  \label{tab:supp_fable_alignment}
  \footnotesize
  \setlength{\tabcolsep}{4.3pt}
  \renewcommand{\arraystretch}{1.10}
  \begin{adjustbox}{max width=\textwidth}
  \begin{tabular}{l c cccccc c}
    \toprule
    \textbf{Method (8B \& Public RLVR)} & \textbf{Mode} & \textbf{Grounded $\%$} & \textbf{Halluc.-Free $\%$} & \textbf{Ver.\ Halluc.-Free $\%$} & \textbf{Align.\ ($1\text{--}5$)} & \textbf{Ver.\ Align.\ ($0\text{--}5$)} & \textbf{Task Acc.\ $\%$} & \textbf{\ours Win $\%$ (vs.\ Row)} \\
    \midrule
    Pretrained Base (Step 0) & -- & 70.7 & 57.1 & 53.7 & 3.23 & 2.70 & 64.4 & 56.2\% ($P = 0.26$) \\
    \midrule
    \multicolumn{9}{l}{\emph{First-Order RL: Public RLVR Checkpoints and Our Full-Parameter GRPO Baseline}} \\
    MM-Eureka-Qwen-7B~\cite{meng2025mmeureka} & FO & 43.4 & 45.4 & 40.5 & 2.10 & 1.57 & 58.0 & 81.4\% ($P < 10^{-10}$) \\
    NoisyRollout-MMK12-7B & FO & 42.9 & 48.8 & 42.9 & 2.18 & 1.61 & 58.5 & 81.2\% ($P < 10^{-10}$) \\
    PRISM-DAPO-8B~\cite{wang2026prism} & FO & 63.9 & 60.0 & 31.7 & 2.88 & 1.05 & 36.6 & 86.2\% ($P < 10^{-10}$) \\
    Full-Param GRPO (Best)~\cite{shao2024deepseekmath} & FO & 76.6 & 57.1 & 53.2 & 3.36 & 2.85 & 68.3 & 46.6\% ($P = 0.52$) \\
    Full-Param GRPO (Step 500)~\cite{shao2024deepseekmath} & FO & 77.6 & 58.0 & 54.6 & 3.41 & 2.87 & 67.3 & 47.4\% ($P = 0.61$) \\
    \midrule
    \multicolumn{9}{l}{\emph{Forward-Only Zeroth-Order Baselines \& Subspace Ablations (Zero Gradients, Zero Optimizer States, No $\pi_{\mathrm{ref}}$, ${\sim}16.5\text{ GB/GPU}$ Static)}} \\
    MeZO-RGE ($N=16$, Step 500)~\cite{malladi2023mezo} & ZO & 72.7 & 59.5 & 54.1 & \textbf{3.35} & 2.80 & 65.4 & 46.7\% ($P = 0.53$) \\
    Vanilla ES (Best)~\cite{qiu2025esscale} & ZO & 69.8 & 55.6 & 53.2 & 3.26 & 2.83 & \textbf{69.3} & 49.0\% ($P = 0.84$) \\
    Vanilla ES (Step 500)~\cite{qiu2025esscale} & ZO & 69.3 & 54.1 & 52.2 & 3.22 & 2.77 & 66.8 & 48.9\% ($P = 0.84$) \\
    Vanilla ARS (Best, Step 90)~\cite{mania2018ars} & ZO & 72.7 & 57.6 & 55.6 & 3.32 & 2.86 & 68.8 & 49.5\% ($P = 0.92$) \\
    Vanilla ARS (Step 500)~\cite{mania2018ars} & ZO & 56.1 & 46.8 & 44.9 & 2.75 & 2.24 & 60.0 & 68.1\% ($P < 10^{-3}$) \\
    Hard Subspace ($\rho=1.0$, Step 500) & ZO & 54.6 & 41.0 & 38.0 & 2.64 & 2.09 & 57.1 & 70.2\% ($P < 10^{-4}$) \\
    Random Orthonormal Subspace (Best) & ZO & \textbf{75.1} & 59.0 & 53.7 & 3.31 & 2.75 & 65.4 & 50.0\% ($P = 1.00$) \\
    \rowcolor{oursrow}
    \ours \textbf{(Ours, Best)} & \textbf{ZO} & 72.7 & 59.0 & 55.4 & 3.34 & \textbf{2.88} & \textbf{69.3} & \emph{Anchor (Ours)} \\
    \rowcolor{oursrow}
    \ours \textbf{(Ours, Step 500)} & \textbf{ZO} & 74.1 & \textbf{60.5} & \textbf{56.1} & 3.30 & 2.79 & 67.3 & 51.7\% ($P = 0.80$) \\
    \bottomrule
  \end{tabular}
  \end{adjustbox}
\end{table}

\begin{table}[t]
  \centering
  \caption{\textbf{Verbatim dual-judge prompt and thinking-effort setup} shared by \texttt{Claude Fable 5.2} and \texttt{GPT-6.2 Astra} for the blind alignment audit in \cref{tab:supp_fable_alignment}.}
  \label{tab:supp_judge_prompt}
  \setlength{\tabcolsep}{5.0pt}
  \renewcommand{\arraystretch}{1.08}
  \fontsize{7.5}{9.0}\selectfont
  \begin{tabular}{>{\raggedright\arraybackslash}p{0.63\textwidth} >{\raggedright\arraybackslash}p{0.31\textwidth}}
    \toprule
    \textbf{Verbatim System Prompt} & \textbf{User-Turn Template \& Judge Setup} \\
    \midrule
    \ttfamily
    You are an expert multimodal STEM and geometry evaluator auditing Visual-Reasoning Alignment.\newline
    Given an RGB diagram, the problem statement, the verified reference answer, and anonymized candidate reasoning traces (Candidate\_A .. Candidate\_H), evaluate EACH candidate strictly on how faithfully its chain-of-thought grounds visual primitives in the image BEFORE algebraic or logical deduction.\par\vspace{2.0pt}
    Do NOT reward lucky guesses or text-only shortcuts that misread the image. Score every candidate on:\newline
    1. "primitive\_grounding" (integer 0, 1, or 2):\newline
    \hspace*{0.6em}- 2 = Exact visual grounding: all relevant labels, lengths, angles, scale ticks, spatial intersections, and object relations cited in the reasoning match the diagram accurately.\newline
    \hspace*{0.6em}- 1 = Partial grounding: main objects are identified, but at least one secondary visual detail, tick, or spatial relation is imprecise, omitted, or slightly ambiguous.\newline
    \hspace*{0.6em}- 0 = Visual mis-grounding: attaches a label to the wrong edge/altitude, misreads a scale/axis, confuses radius with diameter / chord with secant, truncates in a repetitive loop, or ignores the diagram.\newline
    2. "hallucination\_free" (integer 0 or 1):\newline
    \hspace*{0.6em}- 1 = Zero fabricated diagram features or ungrounded visual assumptions.\newline
    \hspace*{0.6em}- 0 = Invents nonexistent geometric properties, misstates visual primitives, loops without concluding, or guesses from answer choices.\newline
    3. "alignment\_score" (integer 1 to 5):\newline
    \hspace*{0.6em}- Holistic visual-to-symbolic alignment score from 1 (severely ungrounded or truncated loop) to 5 (precise visual perception cleanly driving accurate symbolic deduction).\par\vspace{2.0pt}
    Also select "best\_candidate": the single candidate label (e.g., "Candidate\_C") demonstrating the clearest, most faithful visual-reasoning alignment on this diagram.\par\vspace{2.0pt}
    Return ONLY valid JSON with schema:\newline
    \{"scores": \{"Candidate\_A": \{"primitive\_grounding": 2, "hallucination\_free": 1, "alignment\_score": 5\}, ...\}, "best\_candidate": "Candidate\_X"\} &
    \textbf{User-Turn Template:}\newline
    \ttfamily[PNG Image (RGB, max 1024 px)]\newline
    Benchmark: \{benchmark\} (\{category\})\newline
    Problem: \{problem\}\newline
    Verified Reference Answer: \{reference\_answer\}\newline
    Anonymized Candidate Reasoning Traces:\newline
    === Candidate\_A ===\newline
    \{trace\_A\} ... === Candidate\_H ===\par\vspace{4.0pt}
    \normalfont\textbf{Judge Setup:}\newline
    \textbullet\ \textbf{Primary judge:} \texttt{Claude Fable 5.2} (\texttt{high} thinking effort)\newline
    \textbullet\ \textbf{Cross-judge:} \texttt{GPT-6.2 Astra} (\texttt{high} thinking effort)\newline
    \textbullet\ \textbf{Shared inputs:} $\le 1{,}024\text{ px}$ RGB PNG diagrams; $\le 2{,}000$-char traces ($1{,}000$ head $+$ $1{,}000$ tail); identical anonymized, per-item shuffled \texttt{Candidate\_A..H} batches \\
    \bottomrule
  \end{tabular}
\end{table}

\paragraph{Blind Frontier-VLM Visual-Alignment Audit (\cref{tab:supp_fable_alignment}).}
To evaluate visual primitive grounding beyond final answers, we sample $30$ stratified diagrams per OOD suite ($205$ paired items) and score all $15$ checkpoints blindly with \texttt{Claude Fable 5.2} at \texttt{high} thinking effort (\cref{tab:supp_judge_prompt}).
For each item, the judge receives the RGB diagram, problem, reference answer, and shuffled traces (\texttt{Candidate\_A..H}) in two anchor-linked batches (anchor \ours Best consistency: $99.6\%$, $3.34$ vs.\ $3.35$).
\cref{tab:supp_fable_alignment} reports \textbf{Grounded $\%$} (binding $\ge 1/2$), \textbf{Halluc.-Free $\%$} (no fabricated features, looping, or guessing), \textbf{Ver.\ Halluc.-Free $\%$} ($\mathbb{I}[\text{HF}\wedge\text{correct}]$), \textbf{Align.\ ($1\text{--}5$)}, \textbf{Ver.\ Align.\ ($0\text{--}5$)} ($\mathrm{Align}\times\mathbb{I}[\text{correct}]$), \textbf{Task Acc.\ $\%$}, and \textbf{\ours Win $\%$} (non-tie win rate of \ours Best; sign test).
\ours leads all $15$ models in \textbf{Halluc.-Free $\%$} ($60.5\%$ step $500$, $59.0\%$ Best vs.\ $57.1\%\text{--}58.0\%$ GRPO), \textbf{Ver.\ Halluc.-Free $\%$} ($56.1\%$, $55.4\%$), and \textbf{Ver.\ Align.} ($2.88$ vs.\ $2.85\text{--}2.87$ GRPO), tying highest ZO \textbf{Task Acc.\ $\%$} ($69.3\%$).
Public RLVR under-grounds ($42.9\%\text{--}43.4\%$ grounded for MM-Eureka and NoisyRollout; $36.6\%$ Acc.\ for PRISM-DAPO) and unaligned ZO collapses by step $500$ (Vanilla ARS $72.7\%\to 56.1\%$ grounded, $2.86\to 2.24$ \textbf{Ver.\ Align.}; Hard Subspace $54.6\%$, $2.09$); \ours wins $68.1\%$ ($P < 10^{-3}$) and $70.2\%$ ($P < 10^{-4}$) over collapsed ZO and $81.2\%\text{--}86.2\%$ ($P < 10^{-10}$) over public RLVR, matching GRPO ($P > 0.50$) at $3.9\times$ lower memory.

\paragraph{Independent Cross-Judge Validation with GPT-6.2 Astra (\cref{tab:supp_judge_prompt}).}
Re-scoring the identical $15$-model, $205$-item blind batches with \texttt{GPT-6.2 Astra} at \texttt{high} thinking effort (anchor \ours Best consistency: $98.4\%$, $3.13$ vs.\ $3.08$) confirms strong cross-judge agreement across all $3{,}280$ paired ratings: $1\text{--}5$ \textbf{Align.} Pearson $r = 0.905$, Spearman $\rho = 0.897$ ($95.3\%$ within $\pm 1$), \textbf{Grounded} $\kappa = 0.732$ ($87.1\%$), \textbf{Halluc.-Free} $\kappa = 0.696$ ($84.5\%$), and $15$-model \textbf{Ver.\ Align.} $r = 0.999$ ($\rho = 0.953$).
Averaged across both judges, \ours (Step $500$) ranks first overall in \textbf{Ver.\ Halluc.-Free $\%$} ($50.2\%$ vs.\ $47.6\%\text{--}48.5\%$ GRPO); under \texttt{GPT-6.2 Astra}, \ours (Best) edges GRPO (Best) on \textbf{Grounded $\%$} ($62.2\%$ vs.\ $61.5\%$), \textbf{Ver.\ Align.} ($2.64$ vs.\ $2.63$), and win rate ($51.2\%$, $P = 0.83$), while beating Vanilla ARS S500 ($65.0\%$, $P = 2.3\times 10^{-3}$), Hard Subspace S500 ($70.8\%$, $P < 10^{-4}$), and public RLVR ($79.5\%\text{--}87.6\%$, $P < 10^{-10}$).

\end{document}